\documentclass{article}
\usepackage[utf8]{inputenc} 
\usepackage[T1]{fontenc}    
\usepackage{hyperref}       
\usepackage{url}            
\usepackage{booktabs}       
\usepackage{amsfonts}       
\usepackage{nicefrac}       
\usepackage{microtype}      
\usepackage{graphicx}
\usepackage{lipsum} 
\usepackage{mwe}
\usepackage{wrapfig}
\usepackage{adjustbox}
\usepackage{amsmath} 
\usepackage{authblk}
\usepackage[caption=false,font=footnotesize]{subfig}
\usepackage{multirow} 
\usepackage{tcolorbox}
\usepackage{multicol}
\usepackage{tabularx}
\usepackage{arydshln}
\usepackage{bbding}
\usepackage{enumitem}
\usepackage{balance}
\usepackage{algorithm}

\usepackage{mathrsfs}
\usepackage{listings}
\usepackage[noend]{algpseudocode}
\usepackage{amssymb}
\usepackage{pifont}
\usepackage{hyperref}
\usepackage[normalem]{ulem}
\usepackage{tabularx}
\usepackage{tikz}
\usepackage{bm} 
\usepackage{mathtools}
\usepackage{xspace} 
\newcommand{\ie}{i.e.\@\xspace}

\usepackage{float}
\usepackage{comment}
\usepackage{dashrule} 

\begin{document}

\title{TED++: Submanifold-Aware Backdoor Detection via Layerwise Tubular-Neighbourhood Screening}


\author[1]{Nam Le}
\author[2]{Leo Yu Zhang}
\author[1]{Kewen Liao}
\author[2]{Shirui Pan}
\author[1]{Wei Luo\thanks{Corresponding author: wei.luo@deakin.edu.au}}

\affil[1]{School of Information Technology, Deakin University, Australia\\
\texttt{\{s222576762, kewen.liao, wei.luo\}@deakin.edu.au}}
\affil[2]{School of Information and Communication Technology, Griffith University, Australia\\
\texttt{\{leo.zhang, s.pan\}@griffith.edu.au}}



\maketitle
\begin{abstract}
As deep neural networks power increasingly critical applications, stealthy backdoor attacks, where poisoned training inputs trigger malicious model behaviour while appearing benign, pose a severe security risk. Many existing defences are vulnerable when attackers exploit subtle distance-based anomalies or when clean examples are scarce.
To meet this challenge, we introduce TED++, a submanifold-aware framework that effectively detects subtle backdoors that evade existing defences. TED++ begins by constructing a tubular neighbourhood around each class's hidden-feature manifold, estimating its local ``thickness'' from a handful of clean activations. It then applies Locally Adaptive Ranking (LAR) to detect any activation that drifts outside the admissible tube. By aggregating these LAR-adjusted ranks across all layers, TED++ captures how faithfully an input remains on the evolving class submanifolds.
Based on such characteristic ``tube-constrained'' behaviour, TED++ flags inputs whose LAR-based ranking sequences deviate significantly. Extensive experiments are conducted on benchmark datasets and tasks, demonstrating that TED++ achieves state-of-the-art detection performance 
under both adaptive-attack and limited-data scenarios. 
Remarkably, even with only five held-out examples per class, TED++ still delivers near-perfect 
detection, achieving gains of up to 14\% in AUROC over the next-best method.
The code is publicly available at \url{https://github.com/namle-w/TEDpp}.

\end{abstract}

\section{Introduction}
\begin{figure*}[t]
  \centering
  \subfloat[]{
    \includegraphics[width=0.25\textwidth]{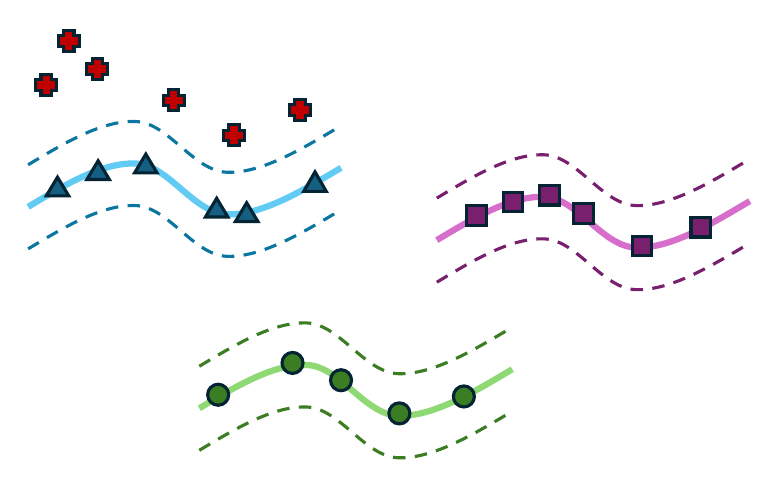}
    \label{fig:combined-concept}
  }
  \hfill
  \subfloat[]{
    \includegraphics[width=0.7\textwidth]{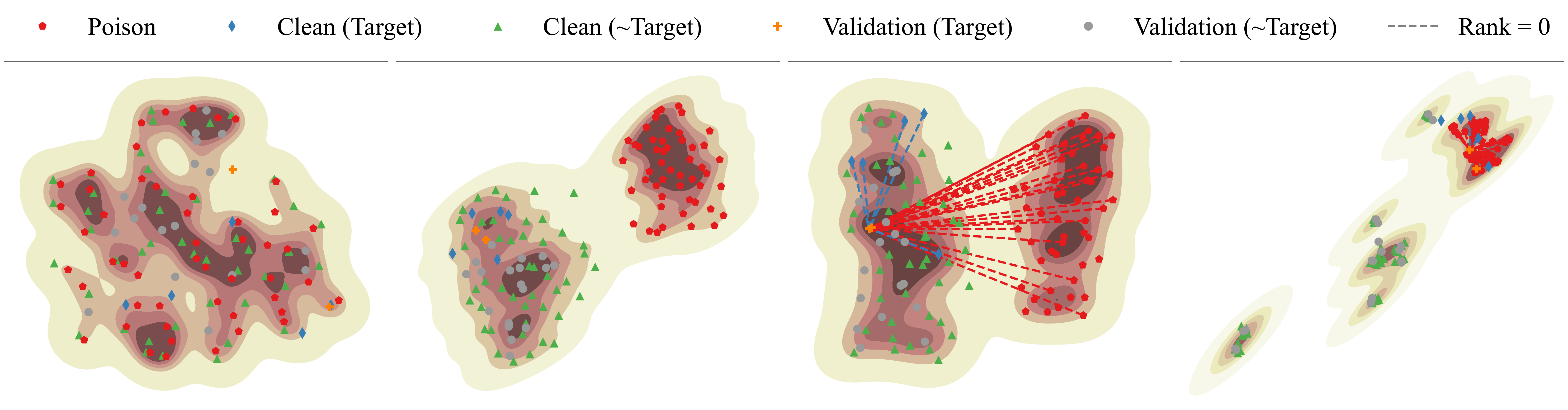}
    \label{fig:combined-umap}
  }
  \caption{%
    \textbf{(a)} Conceptual model of three class submanifolds and their tubular neighbourhoods, with poisoned samples (red) lying just outside the tube yet still having nearest clean neighbours in ambient space.
    \textbf{(b)} 
    UMAP projections of CIFAR-10 activations in four ResNet-18 layers under a Trojan backdoor attack.  
    The shaded contours indicate the estimated tubular neighbourhood around the clean‐class submanifold.
\textbf{Phase 1:} All points—clean and poisoned—lie well inside the tube; ranks are indistinguishable.
\textbf{Phase 2:} Poisoned points begin to drift off the clean manifold (outside the tube), but TED’s rank-based test fluctuates and fails to separate them from clean samples.
\textbf{Phase 3:} Most poisoned samples cross the tube boundary yet still attain the lowest rank 0, showing TED cannot penalise off‐tube departures.
\textbf{Phase 4:} Finally, clean and poisoned trajectories reconverge in the ambient space, rendering TED powerless.
  }
  \label{fig:umap_resnet}
\end{figure*}

Deep neural networks (DNNs) are having a transformative impact on society and have been applied in numerous important domains, such as computer vision \cite{li2022backdoor}. However, their increasing complexity and widespread use have made them vulnerable to significant risks. In particular, backdoor attacks pose a serious threat. In such attacks, an adversary inserts a hidden trigger into the model via poisoned samples during training, causing it to misclassify inputs containing the trigger as target classes, while still correctly classifying clean inputs. These attacks can be devastating, threatening the sustainability and security of the deep learning supply chain, especially when end users have limited control over the training process \cite{guo2020icdm, xu2023icdm}.

According to the application stage in the DNN model lifecycle, defence methods against backdoor attacks currently include data purification \cite{tran2018spectral, li2021anti, jebreel2023defending}, poison suppression \cite{wang2022training, huang2022backdoor, tang2023setting}, model-level detection \cite{wang2019neural, xiang2023umd, wang2024mm} and mitigation \cite{liu2018fine, zeng2022adversarial, guo2023policycleanse}, and input-level detection \cite{gao2021design, liu2023detecting, guo2023scale}. 
Among these, input-level methods have attracted significant interest because they provide the most fine-grained level of security, that is, determining whether an individual input sample is poisoned or benign/clean, while also safeguarding suspicious models that are already deployed.

Previous defences like STRIP \cite{gao2019strip}, TeCo \cite{liu2023detecting}, and SCALE-UP \cite{guo2023scale}  introduce different techniques to transform each image input and monitor how consistent the output is, thereby distinguishing poisoned and clean samples. With the rapid evolution of new backdoor attacks, these defence methods are increasingly insufficient. In response, IBD-PSC~\cite{hou2024ibd} builds on traditional methods by replacing input amplification with activation amplification at multiple batch normalisation layers, allowing it to monitor inconsistencies in model output, as measured by entropy. 
A more recent and robust advancement in this category is Topological Evolution Dynamics (TED) \cite{mo2024robust}. 
Rather than relying on static transformations (e.g., STRIP \cite{gao2019strip}, TeCo \cite{liu2023detecting}, SCALE-UP \cite{guo2023scale}) or metric-based distances computed at a few individual network layers (e.g., IBD-PSC~\cite{hou2024ibd}), TED captures the dynamic evolution of an input as it propagates through the network. It leverages robust topological neighbour relationships and introduces a ranking-based technique, which at each layer assesses a sample’s position as rank $k$ where the sample's $k$-nearest neighbour in a validation set has the sample's target class. 
The core idea is that benign samples
will maintain stable target-class neighbour rankings across all layers, whereas poisoned samples exhibit unstable rankings in the intermediate layers, even though they will ultimately be classified as the target class.

Although TED shows promise in defending against sophisticated backdoor attacks, it
suffers a fundamental breakdown when its test relies on raw rank statistics (based on identifying the top $k$ nearest neighbours) in the full ambient feature space. 
In high-dimensional settings, vast ``empty'' volumes around the low-dimensional data manifold mean that a poisoned activation can drift off-manifold yet still find itself as ``nearest'' neighbour to some distant validation point (See Fig~\ref{fig:umap_resnet}).
The well-known distance concentration phenomenon~\cite{vershynin2018high} further collapses the gap between nearest and farthest distances. 
This pathology is amplified as the validation budget shrinks.
Experimental results highlight this limitation (see Fig~\ref {fig:alpha_blend} and Sec~\ref{sec:exp} for details), indicating that TED fails to detect anomalies effectively and achieves an AUROC score of less than 0.7 when the validation dataset contains fewer than 100 benign samples on CIFAR-10.

To understand why raw rank statistics break down, we traced how clean and poisoned activations evolve through the network. 
By visualising ResNet-18 features at each layer via UMAP---which faithfully preserves both local neighbourhoods and global structure---we see that clean examples tightly hug a low-dimensional submanifold, while poisoned points gradually drift off, only to later reconverge in ambient space (Fig~\ref{fig:umap_resnet}). 
This drift-and-reconverge pattern explains why TED’s ``nearest-neighbour'' rank test fails (particularly with challenging backdoor attacks such as Ada-Patch, Ada-Blend~\cite{qi2023revisiting}, and Trojan~\cite{DBLP:conf/ndss/LiuMALZW018}): without explicit knowledge about the true manifold, off-manifold points can masquerade as neighbours once distances concentrate.

Motivated by these geometric failure modes, we introduce TED++, which exploits the critical information in the submanifold structure by constructing a thin, locally estimated tube around each class’s hidden-feature manifold. By ranking activations relative to that tube boundary---rather than against all ambient-space points---TED++ achieves robust, noise-resilient rank statistics that reliably flag off-manifold (poisoned) inputs even under extreme validation-data scarcity.

We evaluate our method on benchmark datasets CIFAR-10 \cite{krizhevsky2009cifar10}, GTSRB \cite{stallkamp2012man}, and TinyImageNet \cite{le2015tiny} and consistently observe improved performance
over various backdoor attack scenarios,
even with a few validation data points.

\noindent\textbf{1. A submanifold‐centric view of backdoor detection.}
We formalise the intuition that clean‐class activations form low‐dimensional submanifolds in hidden layers.
This insight reveals that existing nearest-neighbour rank defences (e.g.,  TED~\cite{mo2024robust}) ignore this geometric structure and therefore fail to flag subtle, off-submanifold anomalies.
On the other hand, explicit modelling of each submanifold and its tubular neighbourhood~\cite{john2012introduction} yields a robust, noise‐resilient signal for backdoor defence.

\noindent\textbf{2. TED++: Tube‐aware screening with locally adaptive ranking.}
Building on the manifold geometry, we introduce TED++, which (a) estimates a layerwise tubular neighbourhood (or tube) around each class's submanifold, which could be just based on a handful of clean validation examples, (b) applies Locally Adaptive Ranking \textbf{(LAR)} to assign worst‐case ranks to activations lying outside these estimated tubes.
TED++ delivers state-of-the-art backdoor detection performance against an extensive list of attacks.

\section{Related Work}
\subsection{Backdoor Attacks}
Backdoor attacks on neural networks involve adversaries injecting malicious triggers into models to manipulate their behaviour during inference. Generally, existing approaches can be divided into three main categories according to the attackers' capabilities: \textbf{(1)} data poison-only attacks, \textbf{(2)} training-controlled attacks, and \textbf{(3)} model-controlled attacks.

\textbf{Poison-only attacks} refer to scenarios where attackers have no control in the training process but can manipulate the training data by inserting triggers into a part of the dataset. 
A classic approach in this category is BadNets \cite{gu2017badnets}, which takes a small white square as the trigger and superimposes the trigger onto some randomly selected clean images, whose labels are then modified to a specific target class. Training on such poisoned datasets can make models learn a correlation between the injected trigger and the target class. Subsequent research has expanded to attacks that implant more sophisticated triggers that make poisoned samples harder to detect \cite{tang2021demon, qi2023revisiting}, along with clean-label attacks \cite{turner2019label, zeng2023narcissus, gao2023not} where the original labels of the poisoned samples are preserved to avoid detection, and physical attacks that leverage tangible objects or geometric transformations as triggers \cite{wenger2021backdoor, gong2023kaleidoscope}.

\textbf{Training-controlled attacks} allow attackers to manipulate both the training data and the training process. These approaches often attempt to evade detection through sophisticated strategies such as introducing 
``noise modes''~\cite{nguyen2020input, nguyen2021wanet, mo2024robust, zhang2024detector} or special training regularisation techniques~\cite{duan2024conditional}. Another approach in this category focuses on enhancing attack effectiveness \cite{Wangbpp, li2021backdoor, zhang2022poison}, using advanced learning frameworks that go beyond standard supervised learning to embed activations in a sophisticated manner without degrading model performance.

\textbf{Model-controlled attacks} represent scenarios where attackers directly manipulate model architecture or parameters rather than data. An example approach involves injecting additional malicious modules into clean models \cite{tang2020embarrassingly}, while other methods explore parameter-level modifications directly to implant hidden backdoors \cite{qi2022towards}. Though recent studies have explored beneficial uses of backdoors, such as embedding watermarks or protecting intellectual property \cite{li2022untargeted, li2022defending1, li2023black, guo2023domain, tang2023setting, ya2024towards}, such applications are beyond the scope of this work.

In addition to these three capability-based categories, backdoor attacks may also be classified according to (a) the trigger type—whether it is a static \cite{gu2017badnets, chen2017targeted} or a dynamic \cite{nguyen2020input, mo2024robust} pattern that adapts at inference time and (b) the source–target relationship, whether the trigger affects all source classes \cite{DBLP:conf/ndss/LiuMALZW018, nguyen2021wanet} (source-agnostic) or only one or a few specific source classes \cite{mo2024robust, tang2021demon} (source-specific).
Collectively, these various attack strategies demonstrate the growing complexity of backdoor attacks, underscoring the urgent need to develop robust backdoor detection methods capable of adapting to a wide range of adversarial settings.

\begin{figure*}[t]
    \centering
    \includegraphics[width=\textwidth]{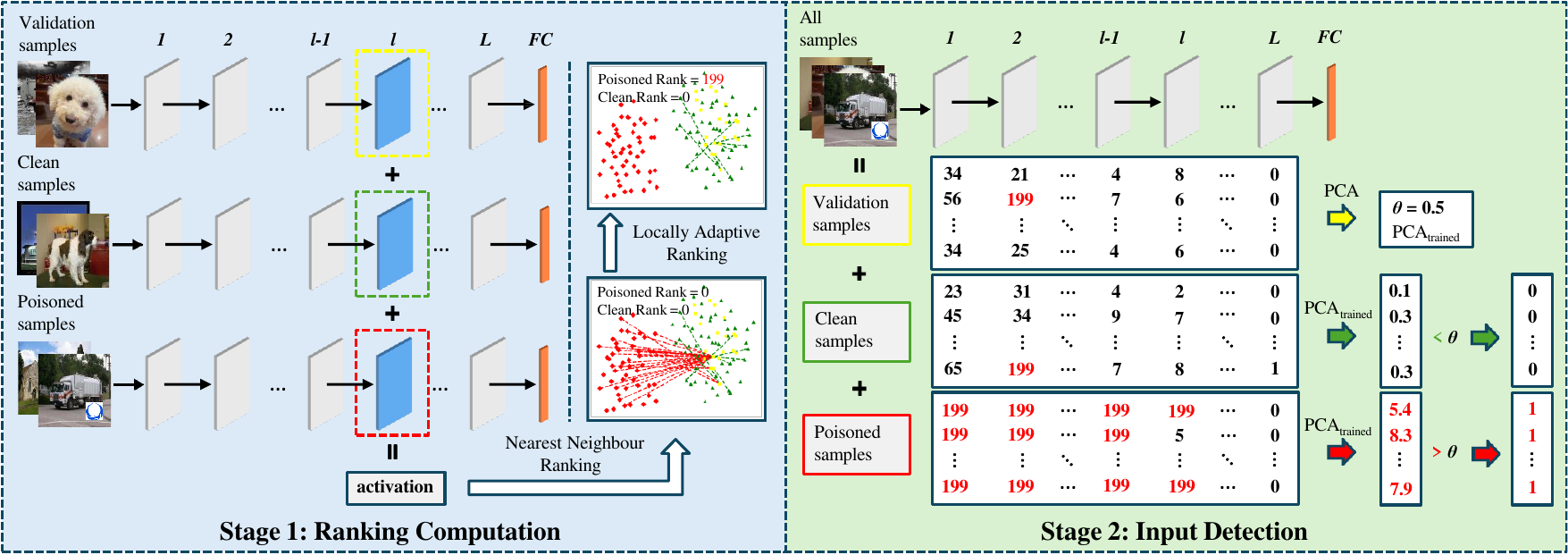} 
    \caption{
TED++ pipeline. 
\textbf{Stage 1 (Ranking Computation):} For each layer \(\ell\), we extract activations \(h^{(\ell)}(x)\), estimate the tubular‐neighbourhood radius \(\tau_\ell\) around each class submanifold using clean validation activations, and apply Locally Adaptive Ranking (LAR) by assigning worst‐case ranks only to off‐tube activations while retaining natural nearest‐neighbour ranks for on‐tube activations. 
\textbf{Stage 2 (Input Detection):} Each sample is represented by its \(L\)-dimensional rank trajectory \(R(x)=[R_1(x),\dots,R_L(x)]\). A PCA model trained on clean‐validation trajectories learns the characteristic ``tube‐constrained'' behaviour; at test time, samples whose reconstruction error exceeds the threshold \(\theta\) are flagged as poisoned.
    }
    \label{fig:diagram}
\end{figure*}
\subsection{Backdoor Defences}
Existing backdoor defence methods can be broadly categorized based on their application stage in the model lifecycle: \textbf{(1)} data purification~\cite{tran2018spectral, li2021anti, jebreel2023defending, 11045703}, \textbf{(2)} poison suppression~\cite{wang2022training, huang2022backdoor, tang2023setting}, \textbf{(3)} model-level detection~\cite{wang2019neural, xiang2023umd, wang2024mm}, \textbf{(4)} model-level mitigation~\cite{liu2018fine, zeng2022adversarial, guo2023policycleanse, zhang2025secure, hou2025fixguard}, and \textbf{(5)} input-level detection~\cite{gao2021design, liu2023detecting, guo2023scale}. 
Data purification techniques aim to remove poisoned samples from training datasets by analysing each sample's influence after model training. Poison suppression approaches alter the training process to keep models from learning backdoor triggers in poisoned samples. Model-level detection approaches often train a meta-classifier or try to replicate activation patterns to evaluate whether a model contains hidden backdoors. Model-level mitigation solutions seek to eliminate or neutralise backdoors built into trained models (see~\cite{mo2025arms} for a comprehensive survey).

Input-level backdoor detection (IBD), the main focus of this paper, functions as a firewall for deployed models by detecting and preventing poisoned inputs at inference time. Compared to other techniques, input-level detection is highly computationally efficient, making it well-suited for scenarios involving third-party models or resource-constrained environments.


STRIP \cite{gao2019strip} is one of the well-known input-level backdoor detection techniques that evaluates the stability of model predictions upon the introduction of benign perturbations to inputs. This approach relies heavily on the assumption that backdoor triggers dominate predictions, ensuring consistency despite perturbations. However, because of this basic idea, STRIP is vulnerable to adaptive backdoor attacks that are specifically built to get around this kind of perturbation-based detection.

In order to detect poisoned inputs, SCALE-UP \cite{guo2023scale}, another well-known IBD technique, amplifies pixel intensities and monitors prediction stability. However, SCALE-UP has natural limitations due to pixel value limits, which limit intensities to a specific range (\ie, [0,255]). Such limitations could unintentionally alter triggered inputs, decreasing the detection accuracy.

More recently, methods like IBD-PSC \cite{hou2024ibd} and TED \cite{mo2024robust} have become the most effective defences against backdoor attacks. Even with few validation samples, IBD-PSC significantly improves detection accuracy and robustness by introducing a parameter-oriented scaling consistency technique. However, despite its strong performance across diverse attacks, it struggles to defend against 
source-specific attacks.
TED is no exception. While it has also shown promise against a variety of attack categories, it suffers from several significant limitations---most notably, its reliance on global nearest ranking information and a large number of validation samples. These limitations are exposed when dealing with sophisticated attacks in practical scenarios. 
Additionally, TED was evaluated on only four image-based attacks, SSDT~\cite{mo2024robust}, IAD~\cite{nguyen2020input}, TaCT~\cite{tang2021demon}, and the Composite Backdoor Attack~\cite{lin2020composite} (a variant of source-specific attacks) in the previous work~\cite{mo2024robust}. Despite the sophistication of these attacks, characterised by dynamic triggers and source-specific behaviour, with triggers that are conspicuous rather than invisible like WaNet \cite{nguyen2021wanet}, TED’s robustness still requires further evaluation.

\section{TED++: Submanifold-Aware Backdoor Detection}
\subsection{Preliminaries}

\textbf{Threat Model.}
This work tackles the problem of detecting input-level backdoors in a white-box setting with limited computing resources. 
Following~\cite{mo2024robust,hou2024ibd}, we consider a white-box DNN $\mathcal{F}: \mathcal{X} \rightarrow [0,1]^C$ with $C$ classes and $L$ hidden layers (standard convolution–BN–ReLU blocks).
We follow the standard poison-only threat model (See \cite{gu2017badnets,mo2024robust}): A small fraction of training samples are augmented with trigger $\delta \coloneq \tau(x)$, yielding a poisoned set $\mathcal{D}_p$.
Defenders have complete access to a suspicious model, but they lack the resources to remove potential backdoors. Following~\cite{mo2024robust}, we assume that defenders have access to a limited set of validation samples with 
at least two validation samples per class.

\textbf{Defenders' Goals.}
An ideal input-level backdoor detection solution should accurately detect and remove all malicious input samples while keeping the model's inference speed intact. In short, defenders aim for two main objectives: \textbf{(1)} Effectiveness---to reliably determine if a suspicious image is malicious, and \textbf{(2)} Efficiency---to 
integrate seamlessly as a plug-and-play module with minimal impact on inference time.

\textbf{Topological Evolution Dynamics in DNNs.}
Topological Evolution Dynamics (TED) \cite{mo2024robust} offers a novel, robust approach to detect backdoor attacks in deep neural networks (DNNs). Unlike traditional methods that assume separability between benign and malicious samples in a fixed metric space, TED views a DNN as a dynamical system evolving inputs into outputs. Benign inputs follow stable evolutionary trajectories within their natural class neighbourhoods across network layers, whereas malicious samples exhibit anomalous trajectories, initially close to benign classes but shifting distinctly towards attacker-specified target classes deeper in the network. TED effectively leverages these unique topological differences to reliably identify backdoored samples across various neural architectures.

\subsection{Motivation and Pipeline Overview }
Our method closely relates to the recently introduced SOTA defence method TED~\cite{mo2024robust}.
TED++’s core novelty is to (i) reframe each class’s clean activations at layer $\ell$ as lying on a smooth submanifold with a learnable ``tube'' of thickness $\tau_l$, and (ii) leverage that tube to detect off-manifold behaviour before ranking.

\textbf{Ambient-Space Neighbour Ranks Overlook Off-Manifold Activations.}
Through the lens of submanifolds,
TED can be interpreted as testing whether a sample’s hidden-layer trajectory crosses one of these class submanifolds, by monitoring nearest-neighbour ranks.  
However, such nearest-neighbour ranks are solely based on ambient-space distances, even when malicious samples lie outside the tubular neighbourhoods (See Fig.~\ref{fig:umap_resnet}).
This limitation makes TED vulnerable to sophisticated backdoor attacks \cite{huang2022backdoor, jebreel2023defending}, where poisoned samples may be significantly distant from benign validation samples in the feature space, yet still appear as nearest neighbours to some target class validation samples in intermediate layers.

\textbf{Two-stage Pipeline.}
To address this limitation, TED++ adopts a two-stage workflow as shown in Fig~\ref{fig:diagram}.
During the \textit{Ranking Computation Stage} (see Sec~\ref{sec:tub} and Sec~\ref{sec:lar}), we explicitly estimate a layer-wise tube radius~$\tau_\ell$ from the clean validation activations, capturing the local "thickness" of each submanifold.  We then introduce Locally Adaptive Ranking (LAR): at each layer, only those activations that lie outside the healthy tube are assigned the worst possible rank, while those inside retain their natural neighbour-order. By aggregating these adjusted ranks across all layers in the \textit{Input Detection Stage} (see Sec~\ref{sec:trajectory}), we focus the detection on genuine backdoor-induced departures from the submanifold rather than benign fluctuations in manifold geometry.  
After collecting the ranking sequence of input $x$ across the considered layers, TED++ leverages a PCA model trained on a benign validation set to predict a final score for input $x$.  
This score is expected to be much higher for poisoned samples and significantly lower for benign ones.  
This submanifold-aware screening endows TED++ with enhanced robustness to small, on-manifold variations while retaining sensitivity to true off-manifold (backdoor) perturbations.  
Alg~\ref{alg:tedpp} illustrates in detail the overall procedure of TED++.

\begin{algorithm}[t]
\caption{The overall algorithm of TED++}
\label{alg:tedpp}
\begin{algorithmic}[1]
\Require 
  Deep network \(f\) with \(L\) layers, \(\{\,\mathcal{V}_c\}_{c=1}^C\) clean validation sets, 
  neighbour‐percentile \(\beta\), PCA model \(\mathrm{PCA}(\cdot)\)
\State \textbf{Preprocessing:}
\For{\(\ell = 1\) to \(L\)}
  \State Compute tube radius \(\tau_\ell\) from \(\{h^{(\ell)}(v):v\in\cup_c\mathcal{V}_c\}\)
\EndFor
\State Fit PCA on \(\{R(v)\}_{v\in\cup_c\mathcal{V}_c}\) and set detection threshold \(\theta\)
\medskip
\State \textbf{Detection:}
\ForAll{\(x\in X_{\mathrm{test}}\)}
  \For{\(\ell = 1\) to \(L\)}
    \State \(z \leftarrow h^{(\ell)}(x)\)
    \State \(v^* \leftarrow \arg\min_{v\in\mathcal{V}_c}\|z - h^{(\ell)}(v)\|_2\)
    \If{\(\|z - h^{(\ell)}(v^*)\|_2 > \tau_\ell\)}
      \State \(R_\ell(x)\leftarrow |\mathcal{V}|\) \Comment{off‐tube $\Rightarrow$ worst rank}
    \Else
      \State \(R_\ell(x)\leftarrow\) rank of \(v^*\) in \(\mathcal{V}_c\) by distance
    \EndIf
  \EndFor
  \State Compute error
  $e \leftarrow \mathrm{PCA}.\,\mathrm{recon\_error}(R(x))$
  \If{\(e > \theta\)} 
    \State Flag \(x\) as poisoned
  \EndIf
\EndFor
\end{algorithmic}
\end{algorithm}
\subsection{Tubular-Neighbourhood Modelling}
\label{sec:tub}

We denote by $h^{(\ell)}(x)\in\mathbb{R}^{d_\ell}$ the activation of input $x$ at layer $\ell$.  Let $\mathcal{C}=\{1,\dots,C\}$ be the set of classes and $\mathcal{V}_c$ a small clean validation set for class $c$.  We make the following assumptions and definitions:

\textbf{Ideal Class Submanifolds.}  
    For each layer $\ell$ and class $c$, we imagine a (smoothly embedded) submanifold
    \begin{equation}
           M_c^{(\ell)} \;\subset\; \mathbb{R}^{d_\ell} 
    \end{equation}
    capturing the true geometry of class-$c$ activations.  These $M_c^{(\ell)}$ are conceptual objects under the manifold hypothesis; different classes may share the same submanifold in some layers.
  
\textbf{Observed Validation Activations.}  
    In practice, we only have access to a finite number of clean validation samples $v\in\mathcal{V}_c$ and their associated activations at layer $l$,  $h^{(\ell)}(v)$, to approximate the geometry of $M_c^{(\ell)}$:   
    \begin{equation}
              H_c^{(\ell)} = \bigl\{ h^{(\ell)}(v) \mid v\in\mathcal{V}_c\bigr\} \subset M_c^{(\ell)}.
    \end{equation}

\textbf{Layer-wise Tubular Neighbourhoods.} 
    Around each submanifold $M_c^{(\ell)}$, define a tubular neighbourhood of radius $\tau_\ell$:
    \begin{equation}
            \mathcal{T}_c^{(\ell)}(\tau_\ell)
      =
      \bigl\{\,z\in\mathbb{R}^{d_\ell}\;\big|\;
        d\bigl(z,\,M_c^{(\ell)}\bigr)\le\tau_\ell
      \bigr\},
    \end{equation}
    where
 $  
      d(z, M) = \inf_{v\in M}\|z - v\|_2.
    $
We choose $\tau_\ell$ so that 
$H_c^{(\ell)} \subset \mathcal{T}_c^{(\ell)}(\tau_\ell).$

By computing each layer’s tubular-neighbourhood radius $\tau_\ell$ as the maximum ``thickness'' of benign activations, we obtain a robust anomaly detector: any input whose layer‐$\ell$ activation falls outside this tube is flagged as suspicious.

Finally, since the ideal submanifold distance is not available in closed form, we approximate each layer–$\ell$ tube radius by measuring the spread of layer–$\ell$ activations of a class \(c\) around a test point \(x\).  Concretely, let \(\mathcal{V}_c\) denote the set of all validation samples of class \(c\).  We locate the layer–$\ell$ \(m\beta\) nearest neighbours of the test activation \(h^{(\ell)}(x)\) in \(\mathcal{V}_c\) and set the tubular radius \(\tau_\ell\) to the maximum of their distances:
\begin{equation}
\max_{1\leq c\leq C}\!\bigl\{d\bigl(h^{(\ell)}(v),h^{(\ell)}(v')\bigr)\bigm|v, v' \in \mathrm{kNN}^{(\ell)}_{\lceil m\beta\rceil}
\bigl(x;\mathcal{V}_c\bigr)\bigr\},
\end{equation}
where $\mathrm{kNN}^{(\ell)}_{m\beta}(x;\,\mathcal{V}_c)$ returns the test point $x$'s \( \lceil m\beta \rceil \) nearest class-$c$ neighbours by Euclidean distance in layer–$\ell$ activations (with \( m \) as the number of validation samples per class and \( \beta \) as the neighbour percentile factor).
This adaptive threshold ensures each tube's ``thickness'' reflects the local density of benign activations in embedding space.

\subsection{Locally Adaptive Ranking}
\label{sec:lar}
Having estimated a layer-wise tube radius $\tau_\ell$ around each class submanifold in the previous section, we now introduce a ranking scheme that explicitly uses this local thickness to detect off-tube activations. Standard nearest-neighbour ranks fail to penalise points that lie off the estimated submanifold, so we propose a \textbf{Locally Adaptive Ranking} (LAR) that assigns the worst possible rank to any activation escaping its tube, while preserving the natural ranking for those that remain inside.

For each input \(x\) with predicted class \(c\), we measure its alignment with the validation manifold at layer \(\ell\) by computing the Euclidean distance between its activation \(h^{(\ell)}(x)\) and the activation of every validation sample \(v\in \mathcal{V}\).  We then sort the full validation set in ascending order of distance:
\begin{equation}
v_{(1)},v_{(2)},\dots,v_{(|
\mathcal{V}|)}
=
\arg\mathrm{sort}_{v\in \mathcal{V}}\;
d\bigl(h^{(\ell)}(x),h^{(\ell)}(v)\bigr),
\end{equation}
where \(v_{(k)}\) is the \(k\)-th nearest neighbour to \(x\) in activation.

Denoting by \(y(v)\) the ground-truth label of \(v\), we define the layer-wise rank of \(x\) as the index of the first neighbour whose true label matches the model’s prediction:
\begin{equation}
R_\ell(x)
\;=\;
\min\bigl\{\,k \;\bigm|\; y\bigl(v_{(k)}\bigr)=c \bigr\}.
\end{equation}
This rank \(R_\ell(x)\) captures how quickly a same-class validation activation appears when we scan the entire set by increasing distance. 

To decide whether $h^{(\ell)}(x)$ lies inside its class-$c$ tube $\mathcal T^{(\ell)}_c(\tau_\ell)$, we first find the closest validation activation of class $c$.  
Specifically, let
\begin{equation}
v^* \;=\;\arg\min_{v\in \mathcal{V}_c}\bigl\|h^{(\ell)}(x)-h^{(\ell)}(v)\bigr\|_2
\end{equation}
be that nearest neighbour.  
We then adjust the rank by checking whether $h^{(\ell)}(x)$ lies outside the tube of radius $\tau_\ell$:
\begin{equation}
R_\ell(x)
=
\begin{cases}
\lvert \mathcal{V}\rvert, 
& \bigl\|h^{(\ell)}(x)-h^{(\ell)}(v^*)\bigr\|_2 > \tau_\ell,\\
R_\ell(x),
& \text{otherwise},
\end{cases}
\end{equation}
where $\lvert \mathcal{V}\rvert$ is the total number of validation samples.  By forcing off-tube activations to take the worst-case rank, LAR sharply increases sensitivity to true off-manifold (backdoor) perturbations while leaving on-tube variations unpenalised.

\subsection{Trajectory Modelling and Detection}
\label{sec:trajectory}
As illustrated by the failure case in Fig.~\ref{fig:umap_resnet}, slight backdoor‐induced perturbations can slip past layer‐wise checks---each individual deviation is too small to trigger an alarm, yet together they form a coherent off‐manifold path.  
Having established in the previous section how Locally Adaptive Ranks quantify per‐layer deviations from the clean‐tube manifold, we now exploit their sequential structure: TED++ models each sample’s rank sequence as a trajectory in $\mathbb{R}^L$ and learns its normal subspace via fitting or training PCA for robust backdoor detection.

Building on the Topological Evolution Dynamics framework~\cite{mo2024robust}, TED++ refines this trajectory‐based subspace approach by incorporating locally adaptive rank statistics.  Concretely, each sample $x\in\mathcal V$ is represented by its sequence of ranks:
\begin{equation}
    R(x) \;=\; \bigl[\,R_1(x),\,R_2(x),\,\dots,\,R_L(x)\bigr],
\end{equation}
where 
\(R_\ell(x)\) is the Locally Adaptive Rank at layer \(\ell\).
Under the \emph{tube‐constrained manifold hypothesis}, clean‐input trajectories occupy a low‐dimensional subspace in $\mathbb{R}^L$, exhibiting only mild, structured fluctuations.
We collect these rank trajectories from all clean validation samples,
\begin{equation}
    \mathcal R = \{\,R(v)\mid v\in\mathcal V\},
\end{equation}
and fit a PCA model to capture their principal modes. 
Let
$
    U_K \in \mathbb{R}^{L\times K}
$
be the matrix whose \(K\) columns are the top \(K\) principal components of \(\mathcal R\).  We then reconstruct any trajectory \(R(x)\) as
\begin{equation}
    \hat R(x) \;=\; U_K\,U_K^\top\,R(x),
\end{equation}
and define its squared reconstruction error
$
    e(x) \;=\; \bigl\lVert R(x) - \hat R(x)\bigr\rVert_2^2.
$
At test time, we declare
\begin{equation}
    \text{Detect}(x) \;=\; \bigl[e(x)>\theta\bigr],
\end{equation}
flagging \(x\) as “poisoned” whenever its trajectory error \(e(x)\) exceeds the threshold \(\theta\) (set to control false alarms on the validation set).

This simple test---driven by the general principle of subspace‐based anomaly detection---efficiently closes the loop from our illustrative failure case in Fig~\ref{fig:umap_resnet} through the layer‐wise rank modelling and into a concrete, low‐overhead implementation that reliably distinguishes true backdoor perturbations from clean variations.

\section{Experiments}\label{sec:exp}
\subsection{Experiment Settings}
\begin{table*}[t]
\centering
\small
\caption{Benign accuracy (BA) and attack success rate (ASR) w.r.t. different attacks and datasets with the ResNet-18 model.}
\label{tab:ba_asr}
\setlength{\tabcolsep}{2pt}
\resizebox{\textwidth}{!}{%
  \begin{tabular}{l*{20}{c}}
    \toprule
    \textbf{Attacks}$\rightarrow$ & \multicolumn{2}{c}{None}
                                & \multicolumn{2}{c}{BadNets}
                                & \multicolumn{2}{c}{Blend} 
                                & \multicolumn{2}{c}{Ada-Patch} 
                                & \multicolumn{2}{c}{Ada-Blend} 
                                & \multicolumn{2}{c}{WaNet} 
                                & \multicolumn{2}{c}{Trojan} 
                                & \multicolumn{2}{c}{IAD} 
                                & \multicolumn{2}{c}{TaCT} 
                                & \multicolumn{2}{c}{SSDT} \\
    \textbf{Datasets}$\downarrow$ & BA & ASR & BA & ASR & BA & ASR & BA & ASR & BA & ASR & BA & ASR & BA & ASR & BA & ASR & BA & ASR & BA & ASR \\
    \midrule
    CIFAR-10 & 0.9391 & ------ & 0.9250 & 1.0000 & 0.9358 & 0.9993 & 0.9375 & 0.8696 & 0.9293 & 0.8693 & 0.9275 & 0.9762 & 0.9349 & 1.0000 & 0.9328 & 0.9999 & 0.9415 & 0.9975 & 0.9404 & 0.9710 \\
    \hline
    GTSRB  & 0.9699 & ------ & 0.9685 & 1.0000 & 0.9707 & 0.9990 & 0.9614 & 0.9984 & 0.9647 & 0.9924 & 0.9596 & 0.9090 & 0.9683 & 1.0000 & 0.9676 & 1.0000 & 0.9714 & 1.0000 & 0.9743 & 0.9888 \\
    \hline
    TinyImageNet  & 0.6081 & ------ & 0.5639 & 0.9328 & ------ & ------ & 0.5836 & 1.0000 & ------ & ------ & ------ & ------ & 0.5759 & 1.0000 & ------ & ------ & ------ & ------ & 0.5948 & 0.9200 \\
    \bottomrule
  \end{tabular}%
}
\end{table*}

\begin{table*}[t]
\centering
\small
\caption{Performance (AUROC, F1) of our method compared to other state-of-the-art defence methods against various attacks on CIFAR-10. TED and TED++ are provided with a small validation set consisting of 5 samples per class. We mark the best result in \textbf{boldface}, the second best result as \underline{underlined} and failed cases ($< 0.7$) in \textcolor{red}{red}.}
\label{tab:CIFAR-10_main}
\setlength{\tabcolsep}{2pt}
\resizebox{\textwidth}{!}{%
  \begin{tabular}{l*{20}{c}}
    \toprule
    \textbf{Attacks}$\rightarrow$ & \multicolumn{2}{c}{BadNets} & \multicolumn{2}{c}{Blend} & \multicolumn{2}{c}{Ada-Patch} & \multicolumn{2}{c}{Ada-Blend} & \multicolumn{2}{c}{WaNet} & \multicolumn{2}{c}{Trojan} & \multicolumn{2}{c}{IAD} & \multicolumn{2}{c}{TaCT} & \multicolumn{2}{c}{SSDT} & \multicolumn{2}{c}{\textit{Avg.}} \\
    \textbf{Defences}$\downarrow$ & AUC & F1 & AUC & F1 & AUC & F1 & AUC & F1 & AUC & F1 & AUC & F1 & AUC & F1 & AUC & F1 & AUC & F1 & AUC & F1 \\
    \midrule
    SCALE-UP     
      & 0.9643 & 0.9151
      & \textcolor{red}{0.6876} & \textcolor{red}{0.5238}
      & 0.7963 & 0.7377
      & 0.7552 & \textcolor{red}{0.6392}
      & 0.7247 & \textcolor{red}{0.6137}
      & 0.9219 & 0.8808
      & 0.9657 & 0.9292 
      & \textcolor{red}{0.6001} & \textcolor{red}{0.2878} 
      & \textcolor{red}{0.4974} & \textcolor{red}{0.1183}
      & 0.7681 & 	\textcolor{red}{0.6273} \\
    STRIP        
      & \textcolor{red}{0.6408} & \textcolor{red}{0.2323} 
      & 0.7389 & \textcolor{red}{0.5608}
      & 0.8212 & \textcolor{red}{0.6847}
      & \underline{0.9103} & \underline{0.8193}
      & \textcolor{red}{0.4564} & \textcolor{red}{0.1131}
      & 0.7115 & \textcolor{red}{0.3099} 
      & 0.9868 & \underline{0.9397} 
      & \textcolor{red}{0.4601} & \textcolor{red}{0.1006} 
      & \textcolor{red}{0.4916} & \textcolor{red}{0.0924}
      & \textcolor{red}{0.6908} & \textcolor{red}{0.4281} \\
    IBD-PSC      
      & \textbf{0.9960} & \textbf{0.9563} 
      & \textbf{0.9950} & \underline{0.9648} 
      & \underline{0.8856} & \underline{0.9166} 
      & 0.8588 & 0.7711 
      & \textbf{0.9736} & \textbf{0.9588} 
      & \textbf{0.9650} & \underline{0.9558} 
      & \textbf{1.0000} & \textbf{0.9726} 
      & \underline{0.8344} & \underline{0.8762} 
      & \textcolor{red}{0.4853} & \textcolor{red}{0.0671}
      & \underline{0.8882} & \underline{0.8266} \\
    TED          
      & 0.9680 & 0.9388
      & \underline{0.9920} & \textbf{0.9709} 
      & 0.8636 & 0.8095 
      & \textcolor{red}{0.6228} & \textcolor{red}{0.0370}
      & \underline{0.9652} & \underline{0.9200}
      & \textcolor{red}{0.6288} & \textcolor{red}{0.1111} 
      & 0.8148 & \textcolor{red}{0.6667} 
      & \textcolor{red}{0.6860} & \textcolor{red}{0.0385}
      & \underline{0.9248} & \underline{0.8478} 
      & 0.8296 & \textcolor{red}{0.6936} \\
    TED++         
      & \underline{0.9944} & \underline{0.9515} 
      & 0.9208 & 0.8283 
      & \textbf{0.9964} & \textbf{0.9709} 
      & \textbf{0.9308} & \textbf{0.8909} 
      & 0.9172 & 0.8788 
      & \underline{0.9424} & \textbf{0.9709} 
      & \underline{0.9988} & 0.9259 
      & \textbf{1.0000} & \textbf{0.9524} 
      & \textbf{0.9980} & \textbf{0.9174} 
      & \textbf{0.9665} & \textbf{0.9581} \\
    \bottomrule
  \end{tabular}%
}
\end{table*}

\begin{table*}[t]
\centering
\small
\caption{Performance (AUROC, F1) of our method compared to other state-of-the-art defence methods against various attacks on GTSRB. TED and TED++ are provided with a small validation set consisting of 5 samples per class. We mark the best result in \textbf{boldface}, the second best result as \underline{underlined} and failed cases ($< 0.7$) in \textcolor{red}{red}.}
\label{tab:gtsrb_main}
\setlength{\tabcolsep}{2pt}
\resizebox{\textwidth}{!}{%
  \begin{tabular}{l*{20}{c}}
    \toprule
    \textbf{Attacks}$\rightarrow$ 
      & \multicolumn{2}{c}{BadNets} 
      & \multicolumn{2}{c}{Blend} 
      & \multicolumn{2}{c}{Ada-Patch} 
      & \multicolumn{2}{c}{Ada-Blend} 
      & \multicolumn{2}{c}{WaNet} 
      & \multicolumn{2}{c}{Trojan} 
      & \multicolumn{2}{c}{IAD} 
      & \multicolumn{2}{c}{TaCT} 
      & \multicolumn{2}{c}{SSDT} 
      & \multicolumn{2}{c}{\textit{Avg.}} \\
    \textbf{Defences}$\downarrow$ 
      & AUC & F1 
      & AUC & F1 
      & AUC & F1 
      & AUC & F1 
      & AUC & F1 
      & AUC & F1 
      & AUC & F1 
      & AUC & F1 
      & AUC & F1 
      & AUC & F1 \\
    \midrule
    SCALE-UP 
      & 0.9025 & 0.8371 
      & \textcolor{red}{0.6224} & \textcolor{red}{0.5577} 
      & 0.8898 & 0.8244 
      & \textcolor{red}{0.5979} & \textcolor{red}{0.5314} 
      & \textcolor{red}{0.3007} & \textcolor{red}{0.1753} 
      & \textcolor{red}{0.2145} & \textcolor{red}{0.0652} 
      & 0.8968 & 0.8321 
      & \textcolor{red}{0.4970} & \textcolor{red}{0.1005} 
      & \textcolor{red}{0.5102} & \textcolor{red}{0.0926}
      & \textcolor{red}{0.6035} & \textcolor{red}{0.4471} \\
      
    STRIP
      & 0.9531 & 0.8885 
      & 0.9154 & \underline{0.8426}
      & \underline{0.9927} & \underline{0.9470} 
      & \underline{0.9390} & \underline{0.8829} 
      & \textcolor{red}{0.4556} & \textcolor{red}{0.1390} 
      & 0.7473 & \textcolor{red}{0.4804} 
      & \textbf{0.9984} & 0.9497 
      & \textcolor{red}{0.4253} & \textcolor{red}{0.0287} 
      & \textcolor{red}{0.5111} & \textcolor{red}{0.1200} 
      & \textcolor{red}{0.7709} & \textcolor{red}{0.5865} \\
      
    IBD-PSC
      & \textbf{0.9638} & \textbf{0.9650} 
      & 0.9116 & \textcolor{red}{0.3605} 
      & 0.9706 & 0.9422 
      & 0.8636 & \textcolor{red}{0.0938} 
      & 0.8855 & \textbf{0.9162} 
      & \textbf{0.9581} & \textbf{0.9525} 
      & 0.9633 & \textbf{0.9642} 
      & \textcolor{red}{0.4876} & \textcolor{red}{0.0000} 
      & \textcolor{red}{0.5312} & \textcolor{red}{0.5303} 
      & 0.8373 & \textcolor{red}{0.6361} \\
      
    TED
      & \underline{0.9566} & \underline{0.9423} 
      & \underline{0.9320} & \textcolor{red}{0.5217} 
      & 0.9432 & 0.9149 
      & 0.7396 & \textcolor{red}{0.6301} 
      & \underline{0.9136} & \underline{0.9091} 
      & 0.8912 & \textcolor{red}{0.5352} 
      & 0.9384 & 0.9320 
      & \underline{0.8440} & \underline{0.7229} 
      & \textbf{0.9996} & \textbf{0.9804} 
      & \underline{0.9065} & \underline{0.8163} \\
      
    TED++
      & 0.9380 & 0.9020 
      & \textbf{0.9964} & \textbf{0.9608} 
      & \textbf{1.0000} & \textbf{0.9709} 
      & \textbf{0.9686} & \textbf{0.9412} 
      & \textbf{0.9138} & 0.8090 
      & \underline{0.9500} & \underline{0.9346} 
      & \underline{0.9764} & \underline{0.9524} 
      & \textbf{0.9166} & \textbf{0.9159} 
      & \underline{0.9576} & \underline{0.8421} 
      & \textbf{0.9575} & \textbf{0.9405} \\
    \bottomrule
  \end{tabular}%
}
\end{table*}

\begin{table}[t]
  \centering
  \small
  \caption{Performance (AUROC, F1) of our method compared to other state-of-the-art defence methods against various attacks on TinyImageNet. TED and TED++ are provided with a small validation set consisting of 5 samples per class. We mark the best result in \textbf{boldface}, the second best result as \underline{underlined} and failed cases ($< 0.7$) in \textcolor{red}{red}.}
  \label{tab:tinyimagenet} 
  \resizebox{\columnwidth}{!}{
  \begin{tabular}{lcccccc} 
    \toprule
    \textbf{Defences}$\rightarrow$ & \multicolumn{2}{c}{IBD-PSC} & \multicolumn{2}{c}{TED} & \multicolumn{2}{c}{TED++} \\
    \textbf{Attacks}$\downarrow$   & AUC     & F1     & AUC     & F1     & AUC     & F1     \\
    \midrule
    BadNets    & \textbf{0.9511} & \textbf{0.9569} & 0.8096  & \textcolor{red}{0.6265}  & \underline{0.8822} & \underline{0.8214} \\
    Ada-Patch  & \textbf{0.9991} & \textbf{0.9795} & \textcolor{red}{0.5848} & \textcolor{red}{0.0000}  & \underline{0.9600} & \underline{0.8772} \\
    Trojan     & \textbf{1.0000} & \textbf{1.0000} & \textcolor{red}{0.6260} & \textcolor{red}{0.0370}  & \underline{0.9340} & \underline{0.8696} \\
    SSDT       & \textcolor{red}{0.3350} & \textcolor{red}{0.0000} & \underline{0.7560} & \textcolor{red}{0.0000}  & \textbf{0.7920} & \textbf{0.7921} \\
    \cdashline{0-6}
    \textit{Average}    & \underline{0.8213} & \underline{0.7341} & \textcolor{red}{0.6934} & \textcolor{red}{0.1659}  & \textbf{0.8920} & \textbf{0.8400} \\
    \bottomrule
  \end{tabular}
  }
\end{table}

\textbf{Datasets and Models.}
We conduct experiments on CIFAR-10 \cite{krizhevsky2009cifar10} and GTSRB \cite{stallkamp2012man} for detailed evaluation, and use TinyImageNet \cite{le2015tiny} as an extension due to computational constraints, employing the ResNet-18 \cite{he2016deep} architecture.                                      

\textbf{Attack Baselines.}
We compare our method to other existing defences against nine representative backdoor attacks: BadNets~\cite{gu2017badnets}, Blend~\cite{chen2017targeted}, Ada-Patch, Ada-Blend~\cite{qi2023revisiting}, TaCT~\cite{tang2021demon}, WaNet~\cite{nguyen2021wanet}, Trojan~\cite{DBLP:conf/ndss/LiuMALZW018}, IAD~\cite{nguyen2020input}, and SSDT~\cite{mo2024robust}. The first five attacks are poison-only attacks, while the remaining four are training-controlled attacks. More specifically, TaCT is a source-specific attack, IAD is a dynamic-trigger attack, and SSDT is both source-specific and dynamic-trigger. All attacks were successfully trained using the ResNet-18~\cite{he2016deep} model, and their corresponding results are presented in Tab~\ref{tab:ba_asr}.

\textbf{Defence Settings.}
We compare our method to state-of-the-art input-level detection defences, including TED~\cite{mo2024robust}, IBD-PSC~\cite{hou2024ibd}, STRIP~\cite{gao2019strip}, and SCALE-UP~\cite{guo2023scale}. We set $\beta=0.5$ to determine the LAR-based adaptive threshold \( \tau_\ell \). Defenders can access 2 to 20 benign samples per class for each dataset for different experiments. 

\textbf{Evaluation Metrics.}
We use two well-recognised metrics: AUROC, which evaluates the performance of detection methods over a range of thresholds; and the F1 score, which integrates both the precision and recall aspects of detection.

\subsection{Main Results}
As shown in Tab~\ref{tab:CIFAR-10_main} and~\ref{tab:gtsrb_main}, TED++ successfully detects poisoned samples, as demonstrated by the AUROC and F1 Score against various attacks.
This makes TED++ robust and comprehensive among current defences, particularly since TED~\cite{mo2024robust} performs poorly with small validation sets and IBD-PSC~\cite{hou2024ibd} cannot detect source-specific attacks such as TaCT~\cite{tang2021demon} and SSDT~\cite{mo2024robust}. 
In addition, SCALE-UP~\cite{guo2023scale} and STRIP~\cite{gao2019strip} are not as effective as the other methods, especially when compared to TED++, despite us using only 5 validation samples per class in all experiments.

After extensive experiments on CIFAR-10~\cite{krizhevsky2009cifar10} and GTSRB~\cite{stallkamp2012man}, we have demonstrated that TED++ is the most effective input-level backdoor defence, as it does not fail against any of the selected attacks. We present additional experiments on a large-scale dataset, involving the top three input-level defences to date against four representative attacks in different categories: BadNets~\cite{gu2017badnets}, Ada-Patch~\cite{qi2023revisiting}, Trojan~\cite{DBLP:conf/ndss/LiuMALZW018}, and SSDT~\cite{mo2024robust}. The experiments are conducted on TinyImageNet~\cite{le2015tiny}, which contains 200 classes. Due to the large number of classes, we aim to make the experiment as realistic as possible by selecting only five samples per class for the validation set for TED and TED++. The results, presented in Tab~\ref{tab:tinyimagenet}, are consistent with previous findings on CIFAR-10 and GTSRB. IBD-PSC performs competitively across all attacks except SSDT, while TED++ maintains strong performance in all attacks, even against SSDT. On the other hand, TED shows the weakest performance due to the validation set limitation. 

\subsection{Ablation Study}
\begin{table*}[t]
  \centering
  \caption{Comparison of AUROC performance between TED and TED++ under various attack types, using different numbers of validation samples per class on CIFAR-10 and GTSRB datasets. We mark better results in \textbf{boldface} and failed cases ($< 0.7$) in \textcolor{red}{red}.}
  \resizebox{\textwidth}{!}{%
    \small
    \begin{tabular}{l*{16}{r}}
      \toprule
      \multicolumn{1}{c}{\multirow{4}{*}{\textbf{Attacks}}} & 
      \multicolumn{8}{c}{\textbf{CIFAR-10}} & 
      \multicolumn{8}{c}{\textbf{GTSRB}} \\
      \cmidrule(lr){2-9} \cmidrule(lr){10-17}
       & \multicolumn{2}{c}{\textbf{20}} & \multicolumn{2}{c}{\textbf{10}} & \multicolumn{2}{c}{\textbf{5}} & \multicolumn{2}{c}{\textbf{2}} &
         \multicolumn{2}{c}{\textbf{20}} & \multicolumn{2}{c}{\textbf{10}} & \multicolumn{2}{c}{\textbf{5}} & \multicolumn{2}{c}{\textbf{2}} \\
      \cmidrule(lr){2-3} \cmidrule(lr){4-5} \cmidrule(lr){6-7} \cmidrule(lr){8-9} 
      \cmidrule(lr){10-11} \cmidrule(lr){12-13} \cmidrule(lr){14-15} \cmidrule(lr){16-17}
       & TED & TED++ & TED & TED++ & TED & TED++ & TED & TED++ &
         TED & TED++ & TED & TED++ & TED & TED++ & TED & TED++ \\
      \midrule
      BadNets    & 0.9548 & \textbf{0.9948} & 0.9724 & \textbf{0.9912} & 0.9680 & \textbf{0.9944} & 0.8380 & \textbf{0.9488} & 
                  0.9604 & \textbf{0.9920} & 0.9546 & \textbf{0.9748} & \textbf{0.9566} & 0.9380 & \textbf{0.9230} & 0.8912 \\
      Blend     & 0.9772 & \textbf{0.9980} & \textbf{0.9844} & 0.9708 & \textbf{0.9920} & 0.9208 & \textcolor{red}{0.3676} & \textbf{0.8836} &
                  0.9864 & \textbf{0.9908} & 0.9620 & \textbf{0.9760} & 0.9320 & \textbf{0.9964} & \textcolor{red}{0.1400} & \textbf{0.8538} \\
      Ada-Patch & 0.8360 & \textbf{0.9928} & 0.8080 & \textbf{0.9924} & 0.8636 & \textbf{0.9964} & \textcolor{red}{0.4548} & \textbf{0.9356} &
                  0.9352 & \textbf{0.9844} & 0.9180 & \textbf{0.9874} & 0.9432 & \textbf{1.0000} & 0.7200 & \textbf{0.9394} \\
      Ada-Blend & 0.7600 & \textbf{0.9984} & \textcolor{red}{0.6356} & \textbf{0.9864} & \textcolor{red}{0.6228} & \textbf{0.9308} & \textcolor{red}{0.6796} & \textbf{0.9616} &
                  0.8980 & \textbf{0.9956} & 0.8556 & \textbf{0.9870} & 0.7396 & \textbf{0.9686} & \textcolor{red}{0.3450} & \textbf{0.8716} \\
      WaNet     & 0.8660 & \textbf{0.9516} & 0.7556 & \textbf{0.9348} & \textbf{0.9652} & 0.9172 & \textbf{0.9180} & 0.8872 &
                  \textbf{0.9246} & 0.9202 & \textbf{0.9400} & 0.8990 & 0.9136 & \textbf{0.9138} & \textbf{0.8928} & 0.8384 \\
      Trojan    & 0.7912 & \textbf{0.9996} & 0.7948 & \textbf{1.0000} & \textcolor{red}{0.6288} & \textbf{0.9424} & 0.7188 & \textbf{0.9660} &
                  0.9462 & \textbf{0.9874} & 0.9370 & \textbf{0.9810} & 0.8912 & \textbf{0.9500} & \textcolor{red}{0.3476} & \textbf{0.8894} \\
      IAD       & 0.8988 & \textbf{0.9956} & 0.8500 & \textbf{0.9944} & 0.8148 & \textbf{0.9988} & \textcolor{red}{0.6132} & \textbf{0.9812} &
                  0.9832 & \textbf{1.0000} & 0.9740 & \textbf{1.0000} & 0.9384 & \textbf{0.9764} & \textbf{0.9996} & 0.9800 \\
      TaCT      & 0.7428 & \textbf{1.0000} & 0.7516 & \textbf{1.0000} & \textcolor{red}{0.6860} & \textbf{1.0000} & 0.8904 & \textbf{1.0000} &
                  0.9328 & \textbf{0.9636} & 0.8916 & \textbf{0.9392} & 0.8440 & \textbf{0.9166} & \textcolor{red}{0.5436} & \textbf{0.8364} \\
      SSDT      & 0.9996 & \textbf{1.0000} & 0.9716 & \textbf{0.9984} & 0.9248 & \textbf{0.9980} & 0.7536 & \textbf{0.9492} &
                  \textbf{0.9996} & 0.9816 & \textbf{0.9996} & 0.9476 & \textbf{0.9996} & 0.9576 & \textbf{0.9388} & 0.8042 \\
      \cdashline{1-17}
      \textit{Average}   & 0.8696 & \textbf{0.9923} & 0.8360 & \textbf{0.9853} & 0.8296 & \textbf{0.9665} & \textcolor{red}{0.6924} & \textbf{0.9459} &
                  0.9518 & \textbf{0.9795} & 0.9369 & \textbf{0.9658} & 0.9065 & \textbf{0.9575} & \textcolor{red}{0.6500} & \textbf{0.8783} \\
      \bottomrule
    \end{tabular}%
  }
  \label{tab:CIFAR-10_gtsrb}
\end{table*}


\begin{figure*}[t]
  \centering
  \subfloat[$m=20$\label{fig:plot1}]{
    \includegraphics[width=.22\linewidth]{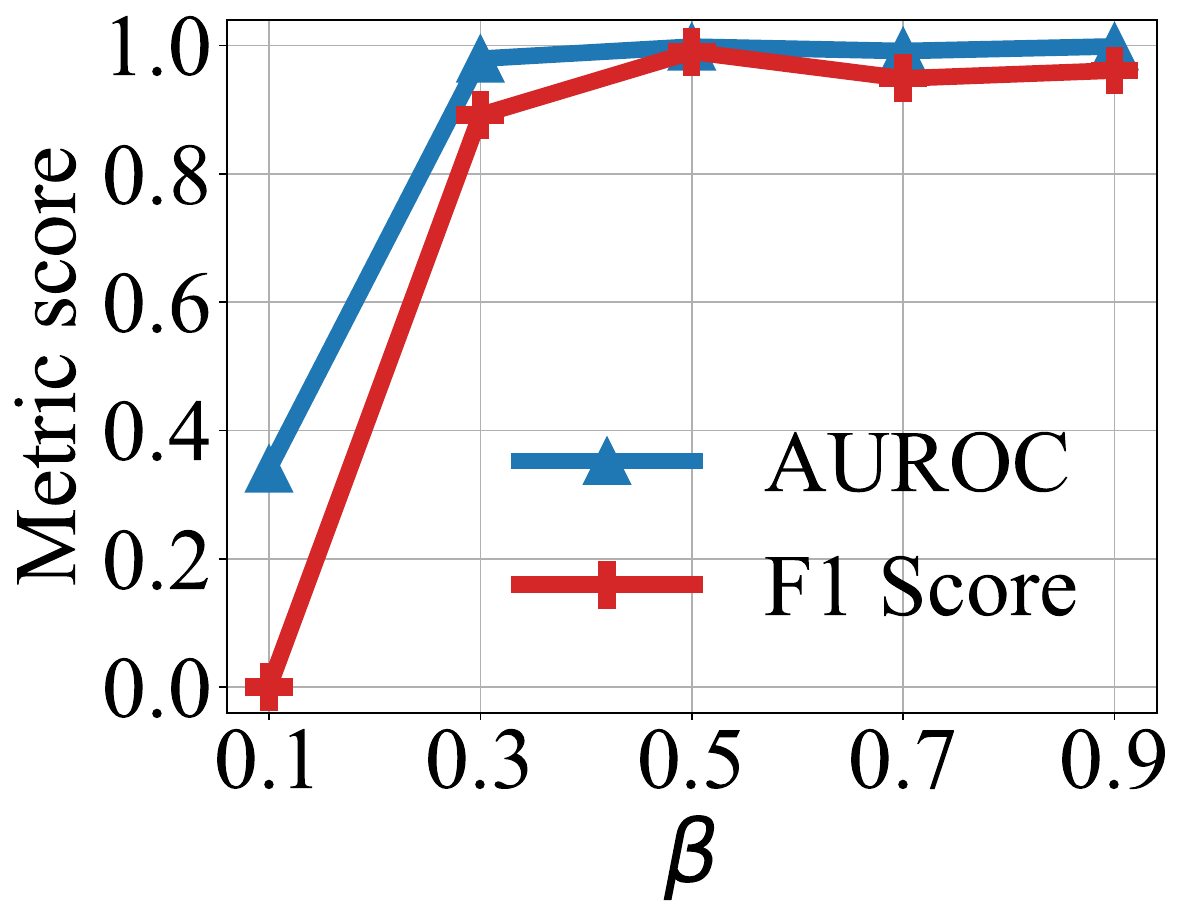}
  }\hfill
  \subfloat[$m=10$\label{fig:plot2}]{
    \includegraphics[width=.22\linewidth]{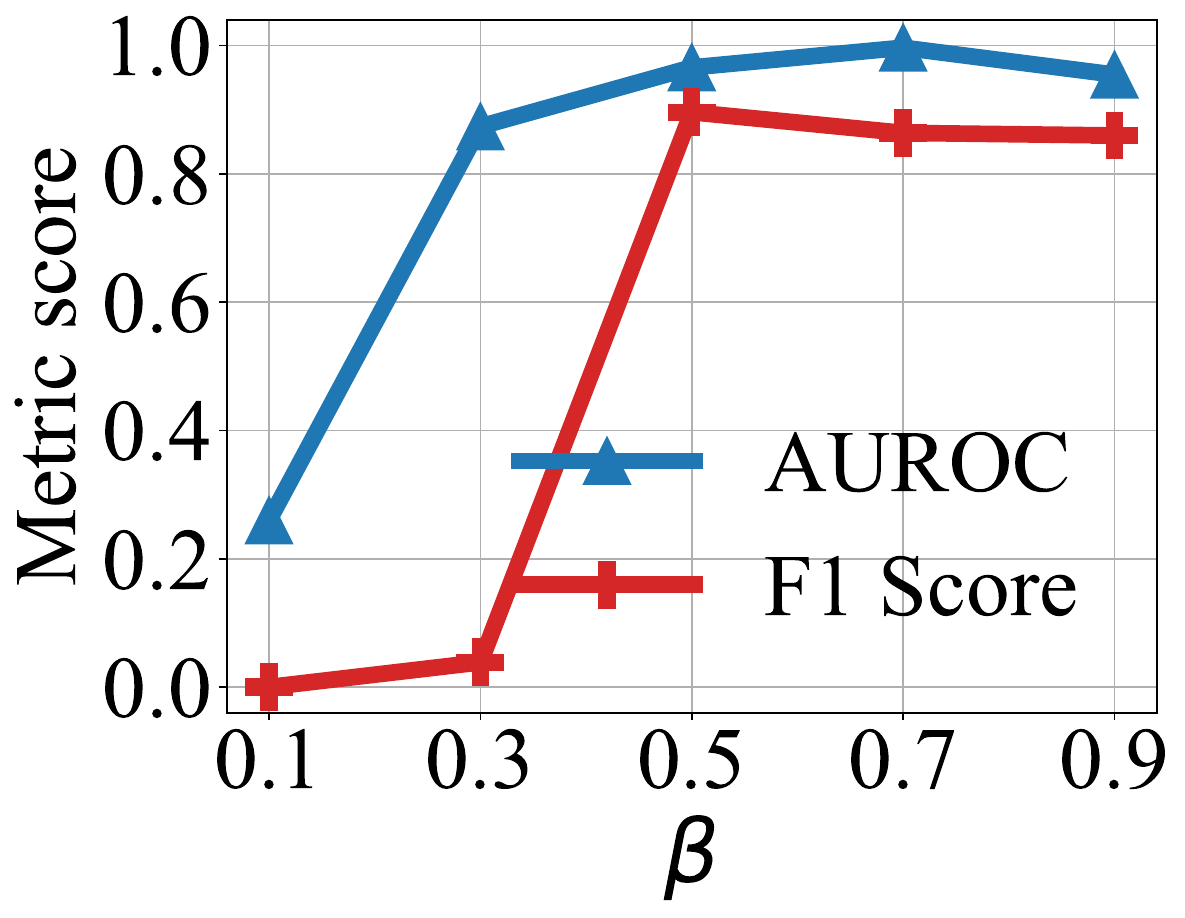}
  }\hfill
  \subfloat[$m=5$\label{fig:plot3}]{
    \includegraphics[width=.22\linewidth]{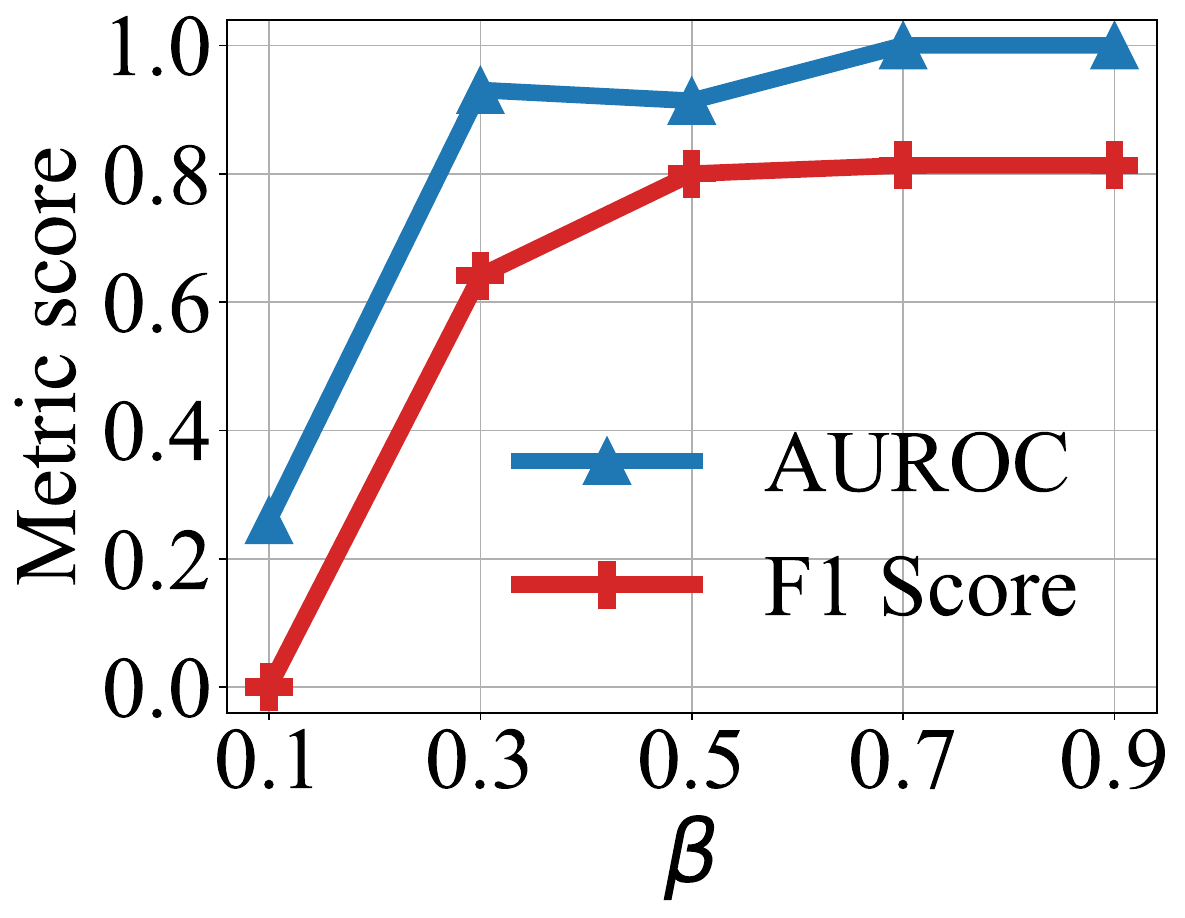}
  }\hfill
  \subfloat[$m=2$\label{fig:plot4}]{
    \includegraphics[width=.22\linewidth]{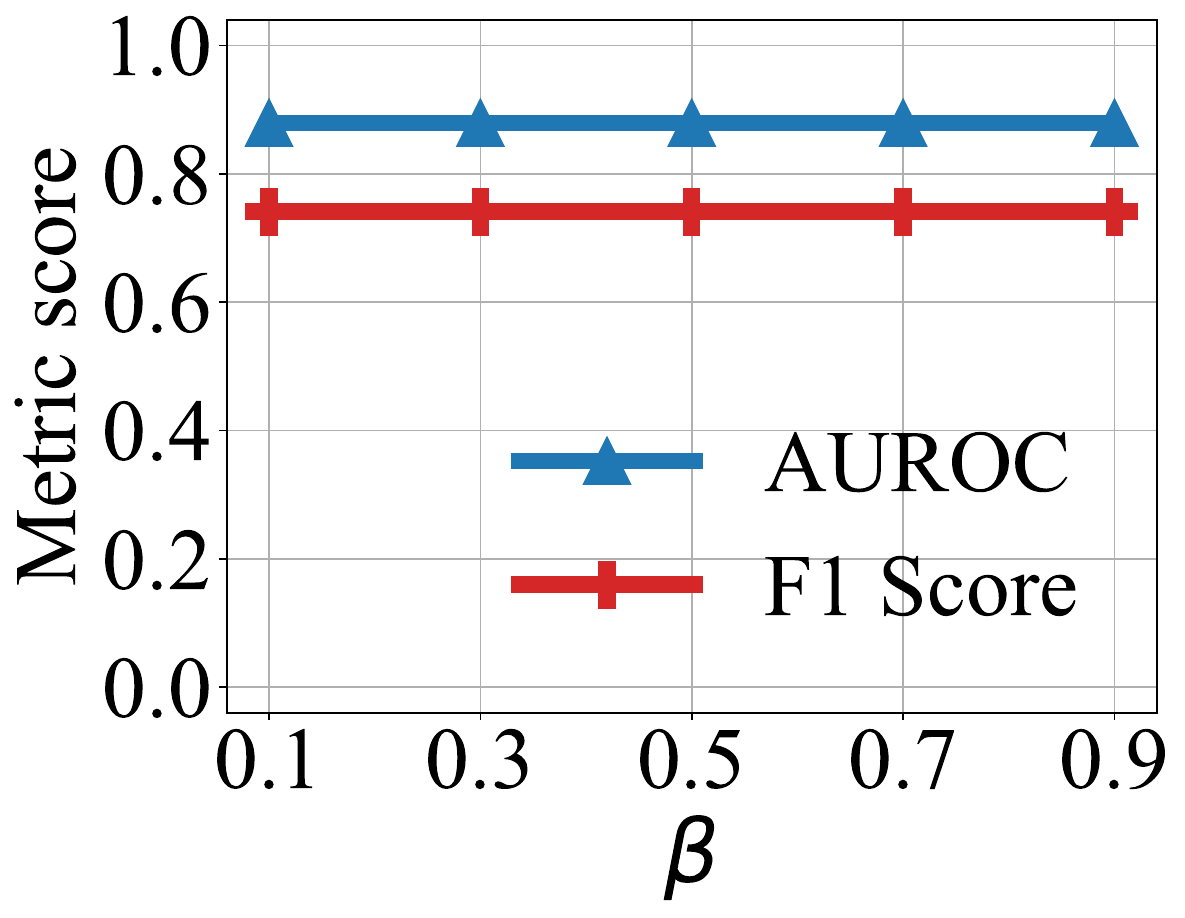}
  }
  \caption{Performance of TED++ against the Blend attack using various combinations of validation sample counts $m$ and the neighbour percentile factor $\beta$ on CIFAR-10.}
  \label{fig:alpha_blend}
\end{figure*}


\begin{figure}[t]
  \centering
  \subfloat[CIFAR-10\label{fig:ted_tedplus:cifar}]{
    \includegraphics[width=.47\linewidth]{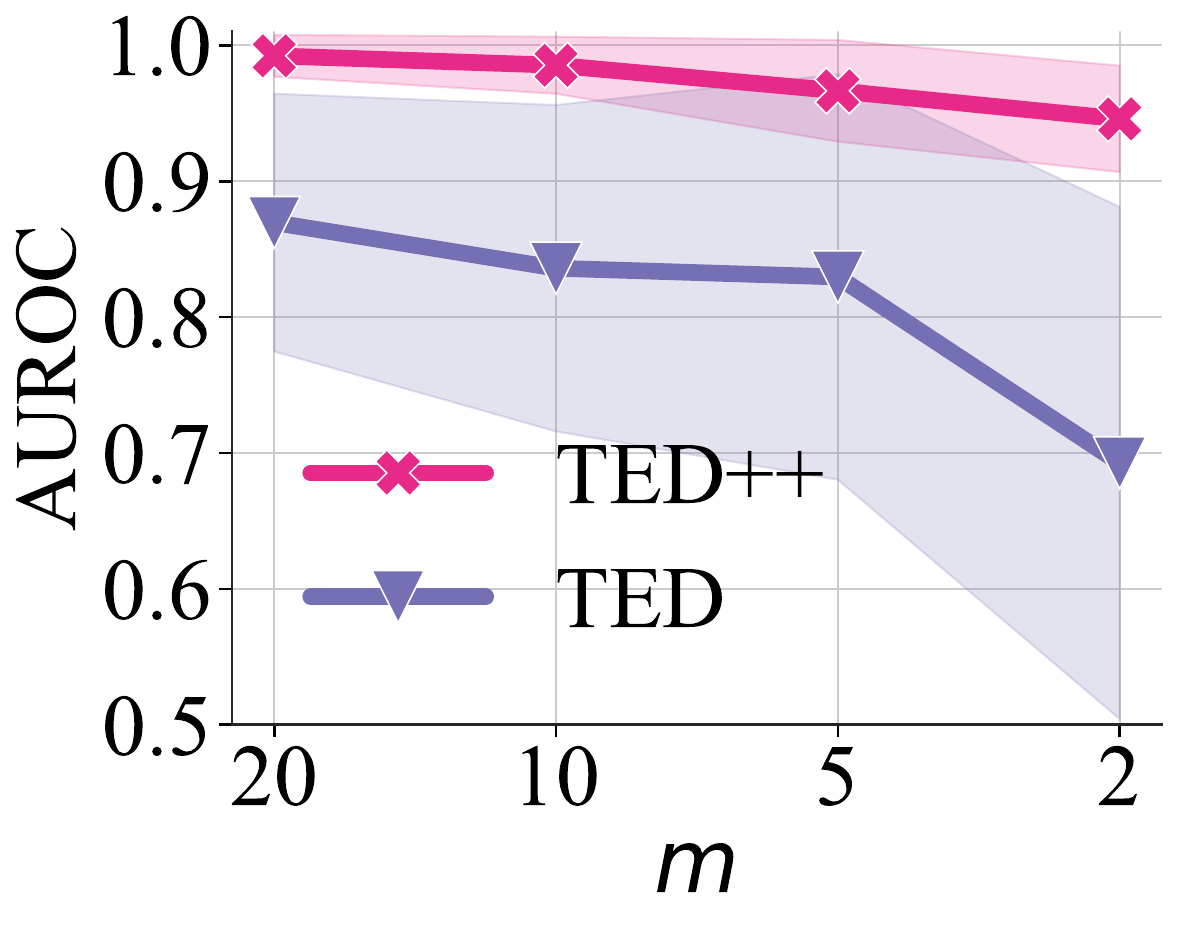}
  }\hfill
  \subfloat[GTSRB\label{fig:ted_tedplus:gtsrb}]{
    \includegraphics[width=.47\linewidth]{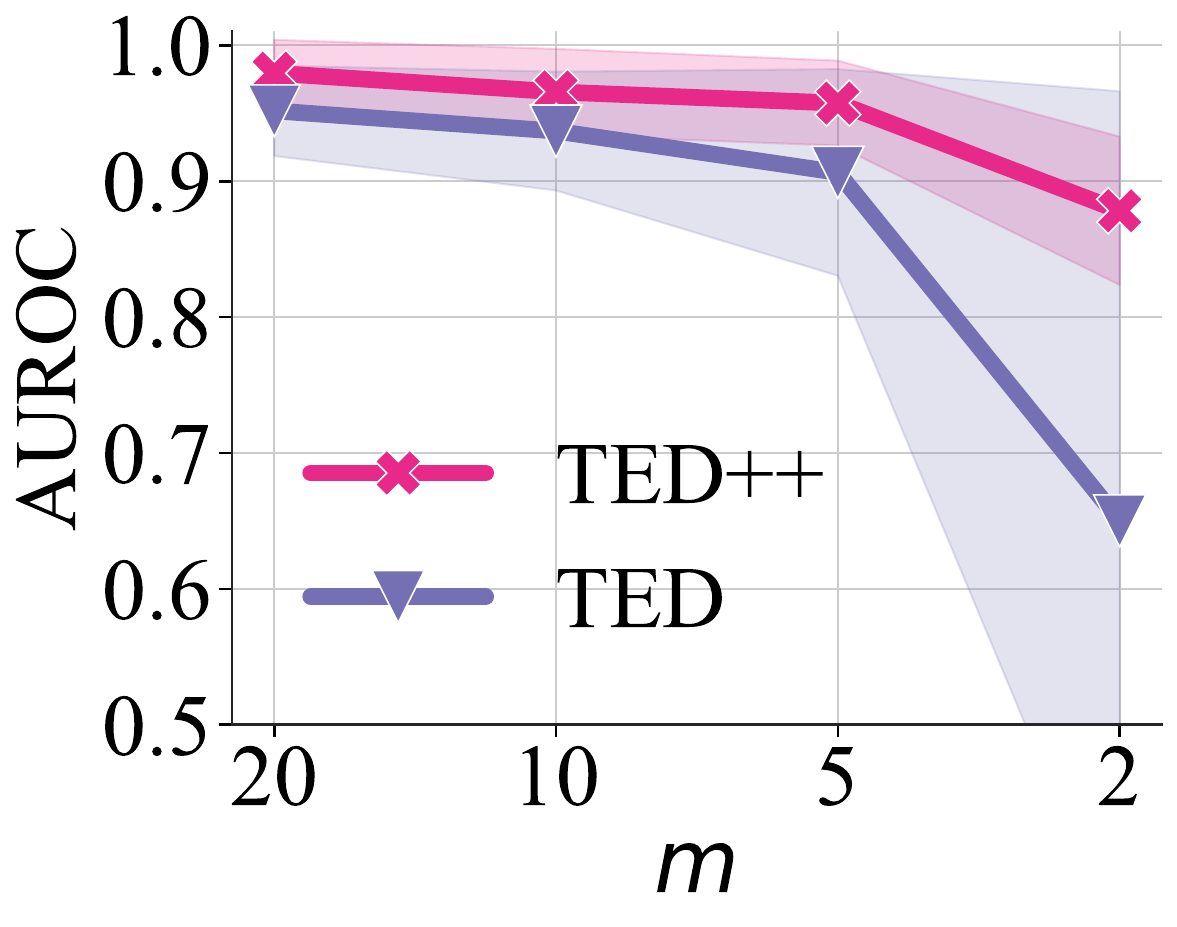}
  }
  \caption{Comparison of AUROC scores between TED and TED++ as the number of validation samples decreases on CIFAR-10 and GTSRB.}
  \label{fig:ted_tedplus}
\end{figure}

\textbf{Impact of the Number of Validation Samples $m$.} For TED++ to operate properly, defenders must be provided with a validation dataset containing at least two samples per class. We investigate the effect of varying the number \(m\) of validation samples per class on the performance of TED++. In particular, we tested TED++ and TED~\cite{mo2024robust} against nine attacks by employing different values of \(m\) ranging from 20 samples per class to 2 samples per class on both CIFAR-10~\cite{krizhevsky2009cifar10} and GTSRB~\cite{stallkamp2012man} in Tab~\ref{tab:CIFAR-10_gtsrb}. The results indicate that TED++ sustains consistent performance even when \(m\) is reduced to its minimum value, whereas the performance of TED deteriorates rapidly, which is illustrated in Fig~\ref{fig:ted_tedplus}. This demonstrates that when validation samples are limited, the specific value of \(m\) has a relatively minor influence on the effectiveness of TED++.

\textbf{Impact of the Neighbour Percentile Factor $\beta$.}
TED++ uses the factor \(\beta\) to determine the number of nearest validation neighbours that establish the threshold for the ranking computation. Based on our observation, each clean sample has a nearest neighbour at a distance comparable to those among other clean samples. In contrast, poisoned samples tend to have a greater distance to their neighbours compared to clean samples. In our algorithm, we assume that if the distance from a sample to its neighbour exceeds the maximum distance between that neighbour and its \(\lceil m\beta \rceil\) nearest neighbours, the suspicious sample will be assigned a high rank, with a selected value being \(\beta = 0.5\). To validate this assumption, we performed experiments, illustrated in Fig~\ref{fig:alpha_blend}, which include line plots for various \(\beta\) values across different numbers of validation samples per class, specifically for TED++ against the Blend~\cite{chen2017targeted} attack on CIFAR-10. The plots show that \(0.9 \geq \beta \geq 0.5\) is optimal because increasing \(\beta\) beyond this range can cause TED++ to fail if there are outliers in the validation set, while lowering it can make the threshold overly sensitive. This finding is further supported by the results presented in Tab~\ref{tab:CIFAR-10_main} and Tab~\ref{tab:gtsrb_main}.

\textbf{Impact of Missing Validation Classes $\rho$.} 
TED++ uses $\rho\in[0,1)$ to quantify the fraction of validation classes missing: $\rho=0$ indicates none are missing, while larger $\rho$ values denote progressively more absent classes. Although TED++ is a robust method performing well against existing backdoor attacks with only a minimal validation dataset, it has a strict requirement: defenders must have at least two validation samples per class for the defence to function properly. This can be particularly challenging, even when defenders have access to a large validation set, as not every class is guaranteed to have two or more samples. The question is whether TED++ can still work when the validation set contains only a subset of the total classes (\ie, when $\rho>0$). In our exploration of how image inputs pass through the network, we observe that the benign samples will be close to each other in the embedding space across different classes, and the distance will only really differ when the input is poisoned. To address this limitation, we apply a technique called \textbf{Nearest-Neighbour Label Flipping}. Specifically, when an input is predicted with a label not present in the validation set, we extract its confidence scores for each label and exclude all labels absent from the validation set. We then flip the input’s predicted label to the remaining class with the highest confidence score. We conduct experiments presented in Tab~\ref{tab:limitation} to evaluate TED++ against diverse attacks on CIFAR-10, GTSRB, and TinyImageNet under varying $\rho$ values. Results indicate that even when $\rho=0.4$, TED++ maintains robust performance. Specifically, the average AUROC scores slightly drop from 0.9665 to 0.8801 on CIFAR-10, from 0.9575 to 0.8694 on GTSRB, and notably increase from 0.8096 to 0.8753 on TinyImageNet as $\rho$ increases from 0 to 0.4. The performance improvement observed on TinyImageNet can be attributed to its large number of classes; reducing the number of classes decreases noise in ranking computations, consequently increasing accuracy. The Nearest-Neighbour Label Flipping procedure is illustrated in Alg~\ref{alg:nnlf}.

\begin{algorithm}[t]
\caption{Nearest-Neighbour Label Flipping}
\label{alg:nnlf}
\begin{algorithmic}[1]
\Require
A $C$-class network $f$; validation sets $\{\mathcal{V}_i\}_{i=1}^{C}$ with
$\mathcal{C}_{\mathrm{val}}\!=\!\{\,i \mid |\mathcal{V}_i|\ge 2\,\}$;  
test example $x$ with confidence vector $\mathbf{p}=f(x)\in[0,1]^C$.
\Statex
\Procedure{NNLF}{$x$}
    \State $\hat{y} \gets \arg\max_{k} \mathbf{p}_k$ \Comment{initial predicted label}
    \If{$\hat{y}\notin\mathcal{C}_{\mathrm{val}}$} \Comment{label absent in validation set}
        \State $\mathbf{p}_{\text{mask}} \gets \mathbf{p}$ with $\mathbf{p}_{k}\!\leftarrow\!0\ \forall\,k\notin\mathcal{C}_{\mathrm{val}}$
        \State $\hat{y} \gets \arg\max_{k\in\mathcal{C}_{\mathrm{val}}} \mathbf{p}_{\text{mask},k}$ \Comment{flip to highest-confidence valid class}
    \EndIf
    \State \Return $\hat{y}$ \Comment{final predicted label}
\EndProcedure
\end{algorithmic}
\end{algorithm}

\begin{table}[t]
  \centering
  \caption{AUROC performance of TED++ under various attack types, with 5 validation samples per class on CIFAR-10 and GTSRB and 2 validation samples per class on TinyImageNet, when $\rho=0.4$ increases from 0\%–40\%. We mark failed cases ($<0.7$) in \textcolor{red}{red}. When $\rho=0.4$, the validation set contains 30 samples for CIFAR-10, 129 samples for GTSRB, and 240 samples for TinyImageNet.}
  \resizebox{\linewidth}{!}{%
    \begin{tabular}{llrrrrr}
      \toprule
      \textbf{Datasets} & \textbf{Attacks} &
      0\% & 10\% &
      20\% & 30\% &
      40\% \\
      \midrule
      \multirow{10}{*}{\centering\textbf{CIFAR-10}}
        & BadNets    & 0.9944 & 0.9442 & 0.9588 & 0.9084 & 0.8656 \\
        & Blend      & 0.9208 & 0.8668 & 0.8960 & 0.8788 & 0.7772 \\
        & Ada-Patch  & 0.9964 & 0.9804 & 0.9952 & 0.9896 & 0.9888 \\
        & Ada-Blend  & 0.9308 & 0.8904 & 0.8872 & 0.9168 & 0.8112 \\
        & WaNet      & 0.9172 & 0.9060 & 0.9192 & 0.8428 & 0.7868 \\
        & Trojan     & 0.9424 & 0.9257 & 0.9329 & 0.8935 & 0.9056 \\
        & IAD        & 0.9988 & 0.9924 & 0.9632 & 0.9848 & 0.9516 \\
        & TaCT       & 1.0000 & 0.9618 & 0.9618 & 0.9430 & 0.9412 \\
        & SSDT       & 0.9980 & 0.9788 & 0.9644 & 0.9260 & 0.8932 \\
        \cdashline{2-7}
        & \textit{Average} & 0.9665 & 0.9385 & 	0.9421 & 0.9204 & 0.8801 \\
      \midrule
      \multirow{10}{*}{\centering\textbf{GTSRB}}
        & BadNets    & 0.9380 & 0.8658 & 0.8156 & 0.8548 & 0.8214 \\
        & Blend      & 0.9964 & 0.9250 & 0.8752 & 0.8868 & 0.9304 \\
        & Ada-Patch  & 1.0000 & 0.9800 & 0.8950 & 0.9828 & 0.9288 \\
        & Ada-Blend  & 0.9686 & 0.9410 & 0.9068 & 0.8954 & 0.8694 \\
        & WaNet      & 0.9138 & 0.8684 & 0.7602 & 0.8247 & 0.8575 \\
        & Trojan     & 0.9500 & 0.9002 & 0.8268 & 0.7896 & 0.7844 \\
        & IAD        & 0.9764 & 1.0000 & 0.9888 & 0.9886 & 1.0000 \\
        & TaCT       & 0.9166 & 0.9102 & 0.8420 & 0.8304 & 0.8032 \\
        & SSDT       & 0.9576 & 0.9248 & 0.8772 & 0.8472 & 0.8296 \\
        \cdashline{2-7}
        & \textit{Average} & 0.9575 & 0.9239 & 0.8653 & 0.8778 & 0.8694 \\
        \midrule
      \multirow{5}{*}{\centering\textbf{TinyImageNet}}
        & BadNets    & 0.8558 & 0.9096 & 0.8780 & 0.9012 & 0.8708 \\
        & Ada-Patch  & 0.9000 & 0.9088 & 0.9436 & 0.9316 & 0.9340 \\
        & Trojan     & 0.7224 & 0.8228 & 0.8824 & 0.8676 & 0.8404 \\
        & SSDT       & 0.7600 & 0.8000 & 0.8400 & 0.8640 & 0.8560 \\
        \cdashline{2-7}
        & \textit{Average} & 0.8096 & 0.8603 & 0.8860 & 0.8911 & 0.8753 \\
      \bottomrule
    \end{tabular}%
  }
  \label{tab:limitation}
\end{table}

\section{Discussion and Future Work}
\textbf{Discussion.}
Fundamentally, TED++ uses locally adaptive ranking and topological evolution dynamics to identify backdoor samples, based on the assumption that the trajectories of poisoned inputs deviate significantly from those of benign samples of the target class. TED++ has consistently demonstrated outstanding performance across a wide range of circumstances during our evaluation. We tested nine representative backdoor attacks on three popular benchmark datasets and achieved outstanding results with AUROC scores above 0.95 in almost all cases. Furthermore, even in challenging circumstances like class imbalance and data scarcity, TED++ maintained exceptional effectiveness, achieving AUROC scores mostly better than 0.8.

Given the robustness of the trajectory-based detection mechanism, 
we find TED++ detects a broad range of backdoor attacks.
As long as an input is poisoned, its trajectory through the neural network inherently differs from that of benign inputs. To date, creating an attack with perfect inseparability in the latent feature space has proven infeasible. Previously, \cite{qi2023revisiting} revisited the assumption of latent separability for backdoor defences and concluded that this assumption can fail by designing Ada-Blend and Ada-Patch attacks. However, in our experiments, these attacks are still easily detected by TED++, demonstrating that this conclusion is not entirely accurate: most layers of the model clearly showcase separability between poisoned and clean samples in the embedding space. However, we acknowledge a practical limitation: when employing extremely deep neural networks with highly complex embedding spaces, the computational cost of rank-based metrics can become high.

\begin{figure}[t]
  \centering
  \includegraphics[width=0.85\linewidth]{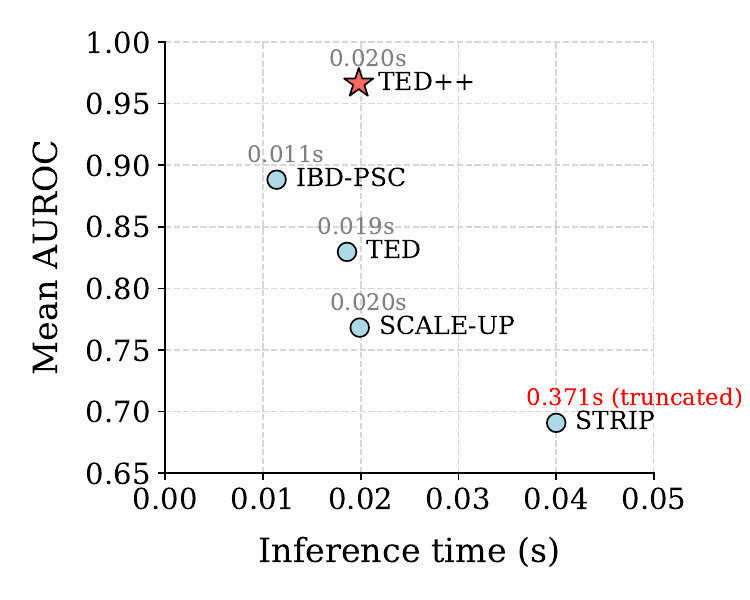}
  \caption{Inference times of state-of-the-art input-level defences on the CIFAR-10 dataset. We provide TED and TED++ with 5 validation samples per class, and apply default settings for the other defences.}
  \label{fig:inference}
\end{figure}

Several critical limitations present in the original TED \cite{mo2024robust} method have been effectively addressed in our work. Specifically, TED’s reliance on a large validation dataset and its expensive computational cost have been significantly alleviated in TED++. Although TED++ still has a comparable computational complexity per inference compared to TED, its dramatically reduced reliance on validation data greatly accelerates the inference process. Fig~\ref{fig:inference} shows the per-sample inference time of all SOTA defences using a 35-layer ResNet-18 model on the CIFAR-10~\cite{krizhevsky2009cifar10} dataset, demonstrating that TED++ achieves inference times comparable to the previous methods while delivering the highest overall performance. These results highlight TED++'s practical suitability for real-time applications.

In comparison with current state-of-the-art input-level backdoor defences, TED++ outperforms all competitors, standing on par only with IBD-PSC~\cite{hou2024ibd}. However, unlike IBD-PSC, TED++ effectively detects source-specific attacks, which remain a critical limitation of IBD-PSC. Hence, we consider TED++ to be the most comprehensive input-level defence solution available to date.

\textbf{Future Work.} 
In this work, we successfully addressed the limitations identified in the previous work~\cite{mo2024robust} by enhancing its capability to detect a diverse set of backdoor attacks. Our experiments position TED++ and IBD-PSC as the two leading input-level defence mechanisms, each with distinct strengths and limitations. While TED++ may encounter computational constraints with very deep networks or intricate embedding spaces, IBD-PSC remains vulnerable to source-specific attacks.

This observation opens a promising direction for future research: designing a next-generation input-level backdoor defence that combines the detection robustness of TED++ with improved computational efficiency. Achieving this goal may require exploring innovative approaches to deep neural network architectures or embedding strategies, which aim to highlight the inherent differences between poisoned and benign inputs without incurring significant computational costs, therefore providing a more scalable and widely applicable solution.

\section{Conclusion}
In this paper, we propose TED++, a robust backdoor defence that outperforms existing methods by effectively detecting backdoor attacks using a minimal validation dataset, a capability that previous defences lack. TED++ was developed based on insights into how poisoned and clean samples propagate through neural networks and interact within the hidden feature space. We observed distinct behavioural patterns of poisoned samples compared to clean ones across most network layers. Recognising the limitations of current defence strategies, we introduced the locally adaptive ranking method in TED++, significantly enhancing detection performance. Specifically, TED++ achieved notable improvements on CIFAR-10~\cite{krizhevsky2009cifar10}, with increases of 16.50\% in AUROC over TED~\cite{mo2024robust}, and 8.12\% in AUROC over IBD-PSC~\cite{hou2024ibd}. On the GTSRB~\cite{stallkamp2012man} dataset, TED++ improved by an average AUROC of 5.63\% compared to TED, and by 14.36\% to IBD-PSC. Moreover, TED++ successfully defended against all tested attacks on TinyImageNet~\cite{le2015tiny}, whereas both IBD-PSC and TED failed. Beyond performance, TED++ addresses critical practical limitations, including reducing computational costs and managing scenarios with insufficient validation classes, ensuring its applicability under diverse conditions. The comprehensive experimental results demonstrate that TED++ is currently the most robust and effective input-level backdoor defence method available.

\bibliographystyle{plain}
\bibliography{arxiv_version} 

\begin{thebibliography}{10}

\bibitem{chen2017targeted}
Xinyun Chen, Chang Liu, Bo~Li, Kimberly Lu, and Dawn Song.
\newblock {Targeted Backdoor Attacks on Deep Learning Systems Using Data Poisoning}.
\newblock {\em arXiv}, 2017.

\bibitem{duan2024conditional}
Qiuyu Duan, Zhongyun Hua, Qing Liao, Yushu Zhang, and Leo~Yu Zhang.
\newblock Conditional backdoor attack via {JPEG} compression.
\newblock In {\em AAAI}, volume~38, 2024.

\bibitem{gao2021design}
Yansong Gao, Yeonjae Kim, Bao~Gia Doan, Zhi Zhang, Gongxuan Zhang, Surya Nepal, Damith~C Ranasinghe, and Hyoungshick Kim.
\newblock Design and evaluation of a multi-domain trojan detection method on deep neural networks.
\newblock {\em IEEE Transactions on Dependable and Secure Computing}, 2021.

\bibitem{gao2019strip}
Yansong Gao, Change Xu, Derui Wang, Shiping Chen, Damith~C Ranasinghe, and Surya Nepal.
\newblock {S}trip: {A} defence against trojan attacks on deep neural networks.
\newblock In {\em ACSAC}, pages 113--125, 2019.

\bibitem{gao2023not}
Yinghua Gao, Yiming Li, Linghui Zhu, Dongxian Wu, Yong Jiang, and Shu-Tao Xia.
\newblock Not all samples are born equal: {T}owards effective clean-label backdoor attacks.
\newblock {\em Pattern Recognition}, 2023.

\bibitem{gong2023kaleidoscope}
Xueluan Gong, Ziyao Wang, Yanjiao Chen, Meng Xue, Qian Wang, and Chao Shen.
\newblock {K}aleidoscope: {P}hysical backdoor attacks against deep neural networks with {RGB} filters.
\newblock {\em IEEE Transactions on Dependable and Secure Computing}, 2023.

\bibitem{gu2017badnets}
Tianyu Gu, Brendan Dolan-Gavitt, and Siddharth Garg.
\newblock Badnets: {I}dentifying vulnerabilities in the machine learning model supply chain.
\newblock {\em arXiv preprint arXiv:1708.06733}, 2017.

\bibitem{guo2023policycleanse}
Junfeng Guo, Ang Li, Lixu Wang, and Cong Liu.
\newblock {PolicyCleanse}: {B}ackdoor detection and mitigation in reinforcement learning.
\newblock In {\em ICCV}, 2023.

\bibitem{guo2023scale}
Junfeng Guo, Yiming Li, Xun Chen, Hanqing Guo, Lichao Sun, and Cong Liu.
\newblock {SCALE-UP}: {A}n efficient black-box input-level backdoor detection via analyzing scaled prediction consistency.
\newblock In {\em ICLR}, 2023.

\bibitem{guo2023domain}
Junfeng Guo, Yiming Li, Lixu Wang, Shu-Tao Xia, Heng Huang, Cong Liu, and Bo~Li.
\newblock Domain watermark: {E}ffective and harmless dataset copyright protection is closed at hand.
\newblock In {\em NeurIPS}, 2023.

\bibitem{guo2020icdm}
Wenbo Guo, Lun Wang, Yan Xu, Xinyu Xing, Min Du, and Dawn Song.
\newblock Towards inspecting and eliminating trojan backdoors in deep neural networks.
\newblock In {\em ICDM}, 2020.

\bibitem{he2016deep}
Kaiming He, Xiangyu Zhang, Shaoqing Ren, and Jian Sun.
\newblock Deep residual learning for image recognition.
\newblock In {\em CVPR}, 2016.

\bibitem{hou2024ibd}
Linshan Hou, Ruili Feng, Zhongyun Hua, Wei Luo, Leo~Yu Zhang, and Yiming Li.
\newblock {IBD-PSC}: {I}nput-level backdoor detection via parameter-oriented scaling consistency.
\newblock In {\em ICML}, 2024.

\bibitem{hou2025fixguard}
Linshan Hou, Zhongyun Hua, Wei Luo, and Leo~Yu Zhang.
\newblock Fixguard: Repairing backdoored models via class-wise trigger recovery and unlearning.
\newblock {\em IEEE Signal Processing Letters}, 2025.

\bibitem{11045703}
Linshan Hou, Wei Luo, Zhongyun Hua, Songhua Chen, Leo Yu~Zhang, and Yiming Li.
\newblock Flare: Toward universal dataset purification against backdoor attacks.
\newblock {\em IEEE Transactions on Information Forensics and Security}, 20:6459--6473, 2025.

\bibitem{huang2022backdoor}
Kunzhe Huang, Yiming Li, Baoyuan Wu, Zhan Qin, and Kui Ren.
\newblock Backdoor defense via decoupling the training process.
\newblock In {\em ICLR}, 2022.

\bibitem{jebreel2023defending}
Najeeb~Moharram Jebreel, Josep Domingo-Ferrer, and Yiming Li.
\newblock Defending against backdoor attacks by layer-wise feature analysis.
\newblock In {\em SIGKDD}, 2023.

\bibitem{krizhevsky2009cifar10}
Alex Krizhevsky.
\newblock The cifar-10 dataset.
\newblock \url{https://www.cs.toronto.edu/~kriz/cifar.html}, 2009.

\bibitem{le2015tiny}
Ya~Le and Xuan Yang.
\newblock Tiny {I}mage{N}et visual recognition challenge.
\newblock {\em CS231n}, 7(7):3, 2015.

\bibitem{john2012introduction}
John~M Lee.
\newblock {\em Introduction to smooth manifolds}.
\newblock Springer, 2012.

\bibitem{li2021anti}
Yige Li, Xixiang Lyu, Nodens Koren, Lingjuan Lyu, Bo~Li, and Xingjun Ma.
\newblock Anti-backdoor learning: {T}raining clean models on poisoned data.
\newblock In {\em NeurIPS}, 2021.

\bibitem{li2022untargeted}
Yiming Li, Yang Bai, Yong Jiang, Yong Yang, Shu-Tao Xia, and Bo~Li.
\newblock Untargeted backdoor watermark: {T}owards harmless and stealthy dataset copyright protection.
\newblock In {\em NeurIPS}, 2022.

\bibitem{li2022backdoor}
Yiming Li, Yong Jiang, Zhifeng Li, and Shu-Tao Xia.
\newblock Backdoor learning: {A} survey.
\newblock {\em IEEE Transactions on Neural Networks and Learning Systems}, 2022.

\bibitem{li2021backdoor}
Yiming Li, Tongqing Zhai, Yong Jiang, Zhifeng Li, and Shu-Tao Xia.
\newblock Backdoor attack in the physical world.
\newblock In {\em ICLR Workshop}, 2021.

\bibitem{li2022defending1}
Yiming Li, Linghui Zhu, Xiaojun Jia, Yong Jiang, Shu-Tao Xia, and Xiaochun Cao.
\newblock Defending against model stealing via verifying embedded external features.
\newblock In {\em AAAI}, 2022.

\bibitem{li2023black}
Yiming Li, Mingyan Zhu, Xue Yang, Yong Jiang, Tao Wei, and Shu-Tao Xia.
\newblock Black-box dataset ownership verification via backdoor watermarking.
\newblock {\em IEEE Transactions on Information Forensics and Security}, 2023.

\bibitem{lin2020composite}
Junyu Lin, Lei Xu, Yingqi Liu, and Xiangyu Zhang.
\newblock Composite backdoor attack for deep neural network by mixing existing benign features.
\newblock In {\em CCS}, pages 113--131, 2020.

\bibitem{liu2018fine}
Kang Liu, Brendan Dolan-Gavitt, and Siddharth Garg.
\newblock Fine-pruning: {D}efending against backdooring attacks on deep neural networks.
\newblock In {\em RAID}, 2018.

\bibitem{liu2023detecting}
Xiaogeng Liu, Minghui Li, Haoyu Wang, Shengshan Hu, Dengpan Ye, Hai Jin, Libing Wu, and Chaowei Xiao.
\newblock Detecting backdoors during the inference stage based on corruption robustness consistency.
\newblock In {\em CVPR}, 2023.

\bibitem{DBLP:conf/ndss/LiuMALZW018}
Yingqi Liu, Shiqing Ma, Yousra Aafer, Wen{-}Chuan Lee, Juan Zhai, Weihang Wang, and Xiangyu Zhang.
\newblock Trojaning attack on neural networks.
\newblock In {\em NDSS}, 2018.

\bibitem{mo2025arms}
Xiaoxing Mo, Nan Sun, Leo~Yu Zhang, Wei Luo, Shang Gao, and Yong Xiang.
\newblock Arms race in deep learning: A survey of backdoor defenses and adaptive attacks.
\newblock In {\em Pacific-Asia Conference on Knowledge Discovery and Data Mining}, pages 308--325. Springer Nature Singapore Singapore, 2025.

\bibitem{mo2024robust}
Xiaoxing Mo, Yechao Zhang, Leo~Yu Zhang, Wei Luo, Nan Sun, Shengshan Hu, Shang Gao, and Yang Xiang.
\newblock Robust backdoor detection for deep learning via topological evolution dynamics.
\newblock In {\em IEEE S\&P}, 2024.

\bibitem{nguyen2020input}
Tuan~Anh Nguyen and Anh Tran.
\newblock Input-aware dynamic backdoor attack.
\newblock In {\em NeurIPS}, 2020.

\bibitem{nguyen2021wanet}
Tuan~Anh Nguyen and Anh~Tuan Tran.
\newblock {WaNet -- Imperceptible Warping-based Backdoor Attack}.
\newblock In {\em ICLR}, 2021.

\bibitem{qi2023revisiting}
Xiangyu Qi, Tinghao Xie, Yiming Li, Saeed Mahloujifar, and Prateek Mittal.
\newblock {Revisiting the Assumption of Latent Separability for Backdoor Defenses}.
\newblock In {\em ICLR}, 2023.

\bibitem{qi2022towards}
Xiangyu Qi, Tinghao Xie, Ruizhe Pan, Jifeng Zhu, Yong Yang, and Kai Bu.
\newblock Towards practical deployment-stage backdoor attack on deep neural networks.
\newblock In {\em CVPR}, 2022.

\bibitem{stallkamp2012man}
Johannes Stallkamp, Marc Schlipsing, Jan Salmen, and Christian Igel.
\newblock Man vs. computer: Benchmarking machine learning algorithms for traffic sign recognition.
\newblock {\em Neural Networks}, 32:323--332, 2012.

\bibitem{tang2021demon}
Di~Tang, XiaoFeng Wang, Haixu Tang, and Kehuan Zhang.
\newblock Demon in the variant: {S}tatistical analysis of dnns for robust backdoor contamination detection.
\newblock In {\em USENIX Security}, 2021.

\bibitem{tang2020embarrassingly}
Ruixiang Tang, Mengnan Du, Ninghao Liu, Fan Yang, and Xia Hu.
\newblock An embarrassingly simple approach for trojan attack in deep neural networks.
\newblock In {\em SIGKDD}, 2020.

\bibitem{tang2023setting}
Ruixiang Tang, Jiayi Yuan, Yiming Li, Zirui Liu, Rui Chen, and Xia Hu.
\newblock Setting the trap: {C}apturing and defeating backdoor threats in plms through honeypots.
\newblock In {\em NeurIPS}, 2023.

\bibitem{tran2018spectral}
Brandon Tran, Jerry Li, and Aleksander Madry.
\newblock Spectral signatures in backdoor attacks.
\newblock In {\em NeurIPS}, 2018.

\bibitem{turner2019label}
Alexander Turner, Dimitris Tsipras, and Aleksander Madry.
\newblock Label-consistent backdoor attacks.
\newblock {\em arXiv}, 2019.

\bibitem{vershynin2018high}
Roman Vershynin.
\newblock {\em High-dimensional probability: An introduction with applications in data science}, volume~47.
\newblock Cambridge university press, 2018.

\bibitem{wang2019neural}
Bolun Wang, Yuanshun Yao, Shawn Shan, Huiying Li, Bimal Viswanath, Haitao Zheng, and Ben~Y Zhao.
\newblock Neural cleanse: {I}dentifying and mitigating backdoor attacks in neural networks.
\newblock In {\em IEEE S\&P}, 2019.

\bibitem{wang2024mm}
Hang Wang, Zhen Xiang, David~J Miller, and George Kesidis.
\newblock {MM-BD}: {P}ost-training detection of backdoor attacks with arbitrary backdoor pattern types using a maximum margin statistic.
\newblock In {\em IEEE S\&P}, 2024.

\bibitem{Wangbpp}
Z.~Wang, J.~Zhai, and S.~Ma.
\newblock {BppAttack}: {S}tealthy and efficient trojan attacks against deep neural networks via image quantization and contrastive adversarial learning.
\newblock In {\em CVPR}, 2022.

\bibitem{wang2022training}
Zhenting Wang, Hailun Ding, Juan Zhai, and Shiqing Ma.
\newblock Training with more confidence: {M}itigating injected and natural backdoors during training.
\newblock In {\em NeurIPS}, 2022.

\bibitem{wenger2021backdoor}
Emily Wenger, Josephine Passananti, Arjun~Nitin Bhagoji, Yuanshun Yao, Haitao Zheng, and Ben~Y Zhao.
\newblock Backdoor attacks against deep learning systems in the physical world.
\newblock In {\em CVPR}, 2021.

\bibitem{xiang2023umd}
Zhen Xiang, Zidi Xiong, and Bo~Li.
\newblock {UMD}: {U}nsupervised model detection for x2x backdoor attacks.
\newblock In {\em ICML}, 2023.

\bibitem{xu2023icdm}
Honghui Xu, Zhipeng Cai, Zuobin Xiong, and Wei Li.
\newblock Backdoor attack on {3D} grey image segmentation.
\newblock In {\em ICDM}, 2023.

\bibitem{ya2024towards}
Mengxi Ya, Yiming Li, Tao Dai, Bin Wang, Yong Jiang, and Shu-Tao Xia.
\newblock Towards faithful xai evaluation via generalization-limited backdoor watermark.
\newblock In {\em ICLR}, 2024.

\bibitem{zeng2022adversarial}
Yi~Zeng, Si~Chen, Won Park, Z~Morley Mao, Ming Jin, and Ruoxi Jia.
\newblock Adversarial unlearning of backdoors via implicit hypergradient.
\newblock In {\em ICLR}, 2022.

\bibitem{zeng2023narcissus}
Yi~Zeng, Minzhou Pan, Hoang~Anh Just, Lingjuan Lyu, Meikang Qiu, and Ruoxi Jia.
\newblock Narcissus: {A} practical clean-label backdoor attack with limited information.
\newblock In {\em CCS}, 2023.

\bibitem{zhang2024detector}
Hangtao Zhang, Shengshan Hu, Yichen Wang, Leo~Yu Zhang, Ziqi Zhou, Xianlong Wang, Yanjun Zhang, and Chao Chen.
\newblock Detector collapse: {B}ackdooring object detection to catastrophic overload or blindness.
\newblock In {\em IJCAI}, 2024.

\bibitem{zhang2022poison}
Jie Zhang, Chen Dongdong, Qidong Huang, Jing Liao, Weiming Zhang, Huamin Feng, Gang Hua, and Nenghai Yu.
\newblock Poison ink: {R}obust and invisible backdoor attack.
\newblock {\em IEEE Transactions on Image Processing}, 2022.

\bibitem{zhang2025secure}
Yechao Zhang, Yuxuan Zhou, Tianyu Li, Minghui Li, Shengshan Hu, Wei Luo, and Leo~Yu Zhang.
\newblock Secure transfer learning: Training clean model against backdoor in pre-trained encoder and downstream dataset.
\newblock In {\em 2025 IEEE Symposium on Security and Privacy (SP)}, pages 1--19. IEEE, 2025.

\end{thebibliography}
\end{document}